\documentclass[10pt,twocolumn,letterpaper]{article}

\usepackage[pagenumbers]{iccv} %

\newcommand{\cP}{\mathcal{P}}

\newcommand{\cI}{\mathcal{I}}
\newcommand{\cX}{\mathcal{X}}

\usepackage{graphicx}
\graphicspath{ {images/} }

\definecolor{vlmcolor}{RGB}{217, 131, 36}
\definecolor{supervisedcolor}{RGB}{107, 142, 35}
\definecolor{sslcolor}{RGB}{74, 119, 184}

\definecolor{augcolor}{RGB}{102, 45, 145}
\definecolor{noaugcolor}{RGB}{0, 180, 180}

\definecolor{lightred}{RGB}{255, 0, 0}   
\definecolor{lightblue}{RGB}{54, 69, 79}
\definecolor{lightgreen}{RGB}{56, 118, 29}

\def\nmsp{\hspace{-6pt}}
\def\nssp{\hspace{-3pt}}
\def\nxssp{\hspace{-1pt}}

\definecolor{appleblue}{RGB}{80,122,255}  %
\definecolor{applepink}{RGB}{217,127,174}  %
\definecolor{appleteal}{RGB}{85,163,152}  %
\definecolor{applepurple}{RGB}{138,107,221}  %
\definecolor{applemustard}{RGB}{234,179,77}  %
\definecolor{appleorange}{RGB}{255,127,80}  %
\definecolor{applegreen}{RGB}{52,199,89}
\definecolor{appleyellow}{RGB}{255,204,0}

\usepackage{tikz}
\usetikzlibrary{arrows,shapes,calc,matrix,fit,backgrounds}
\usepackage{pgfplots}
\usepackage{pgfplotstable}
\pgfplotsset{compat=1.9}
\usepgfplotslibrary{fillbetween}
\usetikzlibrary{patterns}
\usetikzlibrary{calc}
\usetikzlibrary{overlay-beamer-styles}
\usepgfplotslibrary{external}
\usepackage{adjustbox}
\AtBeginEnvironment{tikzpicture}{%
    \renewcommand{\AtTextUpperLeft}{} %
}
\usepackage{algorithm}
\usepackage{algpseudocode}
\usepackage[accsupp]{axessibility}
\usepackage{etoolbox} %

\definecolor{iccvblue}{rgb}{0.21,0.49,0.74}
\usepackage[pagebackref,breaklinks,colorlinks,allcolors=iccvblue]{hyperref}

\def\paperID{10183} %
\def\confName{ICCV}
\def\confYear{2025}

\title{Processing and acquisition traces in visual encoders:\\What does CLIP know about your camera?}

\author{
Ryan Ramos$^{1*}$ \quad Vladan Stojnić$^{2*}$ \quad Giorgos Kordopatis-Zilos$^2$ \quad Yuta Nakashima$^1$  \\ Giorgos Tolias$^2$ \quad Noa Garcia$^1$ \\
$^1$The University of Osaka \quad $^2$VRG, FEE, Czech Technical University in Prague \\
}

\begin{document}

\twocolumn[{%
\renewcommand\twocolumn[1][]{#1}%
\maketitle

\begin{center}
    \includegraphics[width=0.95\textwidth]{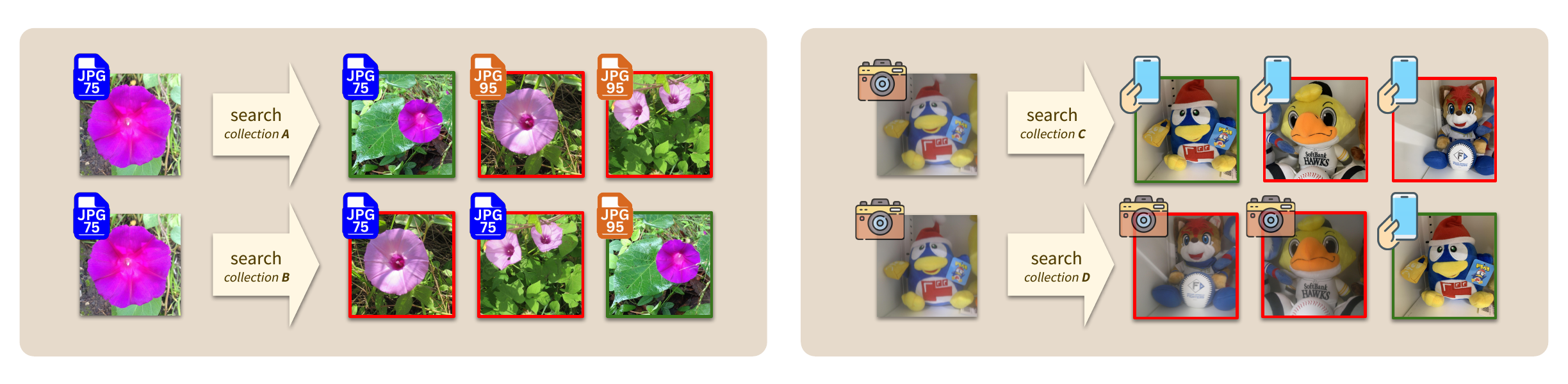}
    \vspace{-10pt}
\captionof{figure}{
\textbf{Impact of metadata on the representation space of visual encoders.} The similarity in the representation space of foundation visual encoders is influenced not only by semantic labels but also by metadata labels. This work explores and uncovers these effects for metadata related to image processing (\eg, JPEG compression) and image acquisition (\eg, camera model). 
The figure illustrates the search results for the query image with specific metadata labels. Retrieved images are ranked according to their similarity to the query. Different image collections exhibit varied combinations of semantic and metadata labels, affecting the retrieval outcome.
}
\label{fig:fig1}
\end{center}%
}]

\maketitle

\def\thefootnote{*}\footnotetext{Equal contribution. 
}
\renewcommand{\thefootnote}{\arabic{footnote}}

\begin{abstract}
Prior work has analyzed the robustness of visual encoders to image transformations and corruptions, particularly in cases where such alterations are not seen during training. When this occurs, they introduce a form of distribution shift at test time, often leading to performance degradation. The primary focus has been on severe corruptions that, when applied aggressively, distort useful signals necessary for accurate semantic predictions.

We take a different perspective by analyzing parameters of the image acquisition process and transformations that may be subtle or even imperceptible to the human eye. We find that such parameters are systematically encoded in the learned visual representations and can be easily recovered. More strikingly, their presence can have a profound impact, either positively or negatively, on semantic predictions. This effect depends on whether there is a strong correlation or anti-correlation between semantic labels 
and these acquisition-based or processing-based labels. Our code and data are available at: \url{https://github.com/ryan-caesar-ramos/visual-encoder-traces}

\end{abstract}

\section{Introduction}
\label{sec:intro}

Visual encoders are fundamental to computer vision algorithms, serving as the backbone for projecting images into meaningful representations. This core pipeline underpins various recognition tasks, from image classification~\cite{krizhevsky2012imagenet,he2016deep} and retrieval~\cite{radenovic2018fine,kordopatis2025ilias} to object detection~\cite{ren2015faster,redmon2016you} and segmentation~\cite{he2017mask,kirillov2023segment}. To ensure broad applicability in downstream tasks, visual encoders are typically pretrained on generic objectives such as supervised image classification~\cite{simonyan2014very,he2016deep}, self-supervised learning~\cite{chen2020simple,caron2021emerging}, or contrastive vision-language training~\cite{radford2021learning,zhai2023sigmoid}. Their primary goal is to capture the semantic content of images, encoding information about objects, attributes, and other relevant features.

Given their central role in computer vision tasks, ensuring the robustness of visual encoders across diverse image distributions is crucial. This has led to extensive research on their sensitivity to various transformations and corruptions, including photometric and geometric changes~\cite{hendrycks2019benchmarking}, noise injection~\cite{hendrycks2019benchmarking}, natural corruptions~\cite{hendrycks2019benchmarking}, image compression~\cite{ghosh2018robustness,chen2023understanding}, adversarial perturbations~\cite{goodfellow2014explaining,kurakin2018adversarial} and examples~\cite{hendrycks2021natural}, and out-of-distribution inputs such as stylized images~\cite{geirhos2018imagenet} or samples from unseen datasets~\cite{hendrycks2021natural}.

This work investigates robustness in the era of foundation models, which are typically kept frozen after large-scale pretraining to preserve their acquired knowledge. However, this practice also preserves any inherent sensitivities or irregularities in their representation spaces. Our study differs from prior work in several key aspects:

\begin{enumerate} \item We examine not only image-processing variations but also acquisition-related factors. We categorize these into two types: \textbf{image processing parameters}, which stem from standard preprocessing and encoding techniques, and \textbf{image acquisition parameters}, which originate from camera settings at capturing time.
Therefore, images carry not only their semantic labels but also \emph{metadata labels} that reflect the values of these acquisition and processing parameters.
\item While existing robustness studies primarily focus on severe corruptions and transformations that degrade the semantic signal, we investigate subtle variations that are often imperceptible to the human eye. 
\item Prior work typically assumes a clean image state during training and evaluates robustness under test-time distribution shifts involving uncommon or degraded states. In contrast, we consider a continuum of valid states, characterized by metadata labels, emphasizing the interplay between the prior distribution of semantic and metadata labels, a key aspect of our findings. \end{enumerate}

To clarify these issues, we address the following questions: \textit{How do 
metadata labels
influence the representations generated by visual encoders?} and \textit{Do
metadata labels
affect the quality of downstream task predictions?}

\begin{figure}[t]
    \centering
    \input{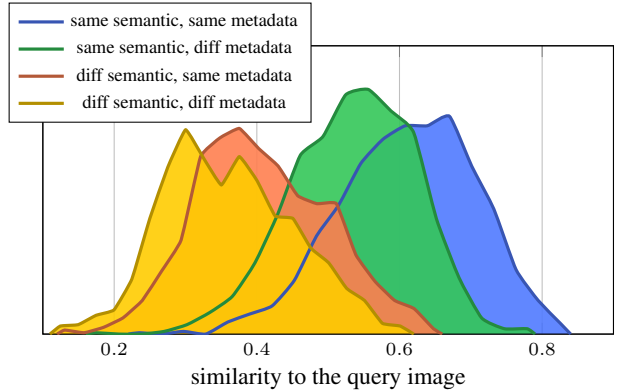}
\pgfplotsset{every tick label/.append style={font=\scriptsize}}
\tikzsetnextfilename{sim_dist}
\begin{tikzpicture}
    \begin{axis}[
        width=1.1\linewidth,
        height=0.65\linewidth,
        label style={font=\footnotesize},
        xmin=0.1, xmax=0.9,
        ymin=0., ymax=0.165,
        ytick=\empty,
        xtick={0.2, 0.4, 0.6, 0.8}, 
        xlabel={similarity to the query image},
        xlabel style={yshift=2pt, font=\small},
        xmajorgrids=true,
        enlargelimits=0.0
    ]
    
        \addplot[ybar, fill=appleblue, smooth,tension=0.3,draw=none, opacity=0.9] table[x=sim, y=dist,each nth point=1] {\possame};

        \addplot[ybar, fill=applegreen, smooth,tension=0.3,draw=none, opacity=0.9] table[x=sim, y=dist,each nth point=1] {\posdiff};

        \addplot[ybar, fill=appleorange,smooth,tension=0.2,draw=none, opacity=0.9] table[x=sim, y=dist,each nth point=1] {\negsame};
        
        \addplot[ybar, fill=appleyellow, smooth,tension=0.2,draw=none, opacity=0.9] table[x=sim, y=dist,each nth point=1] {\negdiff};

    \end{axis}
    \begin{axis}[
        width=1.1\linewidth,
        height=0.65\linewidth,
        xmin=0.1, xmax=0.9,
        ymin=0., ymax=0.165,
        ytick=\empty,
        xtick=\empty,
        enlargelimits=0.0,
        no markers,
        smooth, tension=1.0,
        legend pos=north east,
        xticklabel=\empty,
        legend style={
            font=\scriptsize, 
            at={(-0.06,1.15)},
            anchor=north west, 
            opacity=1.,
        }
    ]
    
        \addplot[appleblue!70!black,smooth,tension=0.3,thick, opacity=1, line width=1.2pt] table[x=sim, y=dist,each nth point=1] {\possame};
        \addlegendentry{same semantic, same metadata}

        \addplot[applegreen!70!black,smooth,tension=0.3,thick, opacity=1, line width=1.2pt] table[x=sim, y=dist,each nth point=1] {\posdiff};
        \addlegendentry{same semantic, diff metadata}

        \addplot[appleorange!70!black,smooth,tension=0.2,thick, opacity=1, line width=1.2pt] table[x=sim, y=dist,each nth point=1] {\negsame};
        \addlegendentry{diff semantic, same metadata}

        \addplot[appleyellow!70!black,smooth,tension=0.2,thick, opacity=1, line width=1.2pt] table[x=sim, y=dist,each nth point=1] {\negdiff};
        \addlegendentry{diff semantic, diff metadata}

    \end{axis}
\end{tikzpicture}
    \vspace{-30pt}
    \caption{\textbf{Distribution of similarities with respect to a query image.}
    Four distributions are shown, based on whether the semantic and metadata labels of the images match those of the query.  
    Metadata labels are based on JPEG quality. The results highlight that similarity is influenced by both types of labels.
    \label{fig:sim_dist}
    }
\end{figure}

Our analysis spans a diverse range of visual encoders, covering different architectures, model sizes, training objectives, and datasets. Key findings include:

\begin{enumerate} \item Acquisition and processing parameters leave identifiable traces in many models’ representation spaces. Such traces allow the inversion, \ie prediction, of the original input image's parameters with considerably higher accuracy than random chance.
\item These traces can overshadow semantic content in downstream tasks, leading to biases where model predictions are influenced by the acquisition or processing parameters rather than the actual object semantics (see \cref{fig:fig1}). 
Interestingly, we find that similarity in the representation space depends on both semantic and metadata labels (see \cref{fig:sim_dist}), with the latter often having a disruptive influence in downstream tasks.
\item We identify contrastive vision-language models as the most sensitive to such influences, and our experiments suggest that this sensitivity stems from the absence of strong image augmentations during pretraining.
\end{enumerate}

These findings raise concerns about model %
reliability, especially in corner cases that, while artificially injected to our experiments, may occur in real-world applications.

\section{Preliminaries}

\paragraph{Visual encoders}
We analyze $47$ visual encoders, each
used to extract image embeddings without any further fine-tuning. We categorize them into three groups according to their training schemes:

\begin{itemize}
    \item \textbf{Supervised} (SUP) are models trained with supervision for image classification on ImageNet~\cite{russakovsky15imagenet}. Several variants of convolutional neural networks (CNN) and Transformer architectures are considered, including ResNet~\cite{he2016deep}, ConvNeXt~\cite{liu2022convnet}, and ViT~\cite{dosovitskiy2020image}. 
    \item \textbf{Self-supervised learning} (SSL) are models trained with contrastive learning. Several variants of DINO~\cite{caron2021emerging}, DINOv2~\cite{oquab2024dinov2}, and MoCov3~\cite{chen2021empirical} are considered.
    \item \textbf{Contrastive visual-language} (CVL) are models trained for vision-language alignment. We use several variants of CLIP~\cite{radford2021learning}, OpenCLIP~\cite{cherti2023reproducible}, and SigLIP~\cite{zhai2023sigmoid,tschannen2025siglip}.
\end{itemize}

\noindent
The full list of visual encoders and their settings can be found in the supplementary material.

    \paragraph{Image processing parameters} are the properties of an image that are not determined at the time of capture and that can be changed in post-processing stages. These include image compression (\eg JPEG quality), color-based transformations (\eg sharpening), image resizing (\eg type of interpolation), \etc. We use the notation $\cP = \left\{p_1,\cdots, p_M\right\}$ to indicate the $M$ possible classes for one of four processing parameters: JPEG compression, sharpening, amount of resizing, and resize interpolation.

    \paragraph{Image acquisition parameters} are the properties of an image that are determined by the method of its capture. These include specification of the camera (\eg camera maker or model) and camera settings used during capturing (\eg exposure, aperture). We use the notation $\mathcal{A} =\{a_1, \dots a_M\}$ to indicate the $M$ possible classes for one of eight acquisition parameters: make, model (all), model (smart), model (smart vs non-smart), exposure, aperture, ISO speed, and focal length.

\begin{table}[t]
\footnotesize
\centering
\vspace{-5pt}
\caption{Image processing (proc.) and image acquisition (acq.) parameters under analysis and their number of classes.}

\setlength{\tabcolsep}{3pt}
\scalebox{0.96}{
\begin{tabular}{p{0.16\textwidth} l r p{0.22\textwidth}}
\toprule
parameter & type & class & description \\
\midrule 

JPEG compression & proc. & 6 & amount of JPEG compression \\ 
sharpening & proc. & 3 & amount of sharpening \\
resizing & proc. & 3 & amount of resizing \\ 
interpolation & proc. & 4 & type of interpolation during resize \\

\midrule

make & acq. & 9 & manufacturer of the camera \\
model {\scriptsize(all)} & acq. & 88 & specific camera model used\\
model {\scriptsize(smart)} & acq. & 12 & specific smartphone used\\
model {\scriptsize(smart vs non-smart)} & acq. & 2 & whether camera is a smartphone\\
exposure & acq. & 16 & amount of light captured by sensor\\
aperture & acq. & 17 & size of the opening in the lens \\
ISO speed & acq. & 16 & camera sensor's sensitivity to light \\
focal length & acq. & 13 & distance from lens to sensor\\

\bottomrule
\end{tabular}
}
\label{tab:parameters}
\end{table}

\cref{tab:parameters} shows the list of parameters under analysis. More details are provided in the supplementary material. %

\paragraph{Datasets}
Image processing parameters can be controlled and adjusted as desired, allowing us to study them on any existing dataset.\footnote{Images in existing datasets are already processed, so our processing is applied on top of any prior one. %
} We use two common image classification datasets that carry semantic labels: ImageNet~\cite{russakovsky15imagenet} and iNaturalist 2018~\cite{horn18inaturalist,inaturalist2018},
following the default splits into training ($\cX_{\text{train}}$) and test ($\cX_{\text{test}}$) sets. In contrast, image acquisition parameters are fixed once an image is captured. They can typically be recovered from the image's Exif tags, which most datasets do not include. We collect and make available two new datasets: FlickrExif and PairCams.

\begin{figure}[t]
    \centering
    \vspace{-10pt}
    \includegraphics[width=0.9\linewidth]{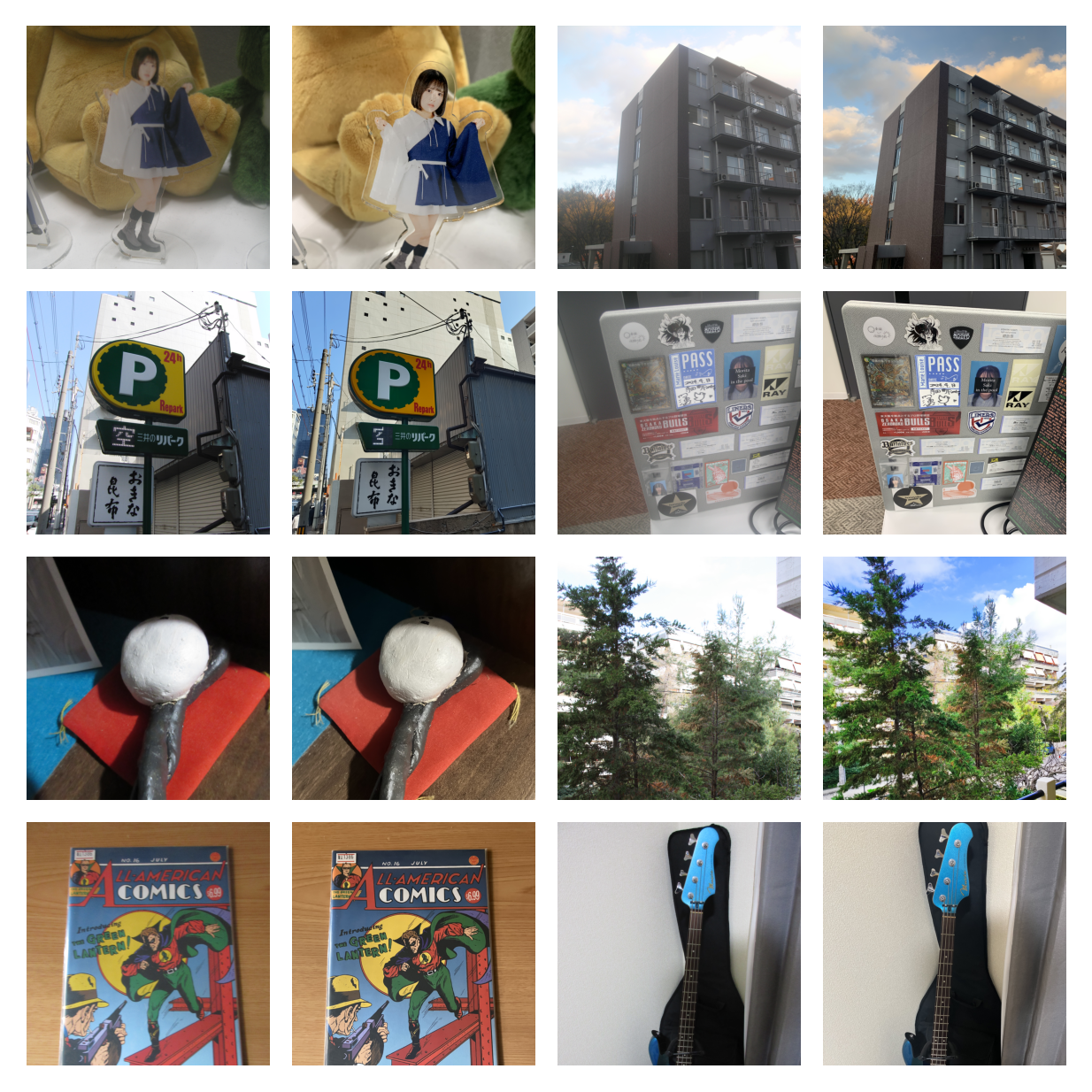}
    \caption{\textbf{Examples of images in the PairCams dataset.} Each pair depicts the same object and/or scene but taken with two different camera types. For each pair, the left image corresponds to a non-smartphone, and the right one to a smartphone.}
  \label{fig:toy_data}
\end{figure}

\begin{itemize}
    \item \textbf{FlickrExif}: We use the Flickr API\footnote{\url{https://www.flickr.com/services/api/}} to download images and their accompanying Exif metadata. For each month from January 2000 to August 2024, we collect between $2,000$ and $4,000$ safe-for-work photos with permissive licenses. To prevent the dataset from being dominated by a small group of prolific photographers, we limit the number of images contributed by each user to 10 per month and year. This leads to a total of $356,459$ images.

    \item \textbf{PairCams}: To ensure complete control over the data and eliminate potential semantic confounders, we manually collect a dataset with $730$ pairs of photos, totaling $1,460$ images. Each pair depicts the same object or scene captured under identical conditions but different \emph{camera types}. Specifically, we use two distinct categories of cameras: (i) modern smartphones, and (ii) older digital cameras.\footnote{All smartphones in PairCams were released in 2018 or later. All digital cameras were released in 2014 or earlier.} All images are taken from the same angle, without flash, with automatic white balancing, automatic exposure mode, and Program AE as the exposure program. \cref{fig:toy_data} shows examples of the collected pairs. The detailed list of cameras used for the collection is provided in the supplementary material. %

\end{itemize}

\begin{figure*}
    \centering
    \vspace{-10pt}
    \input{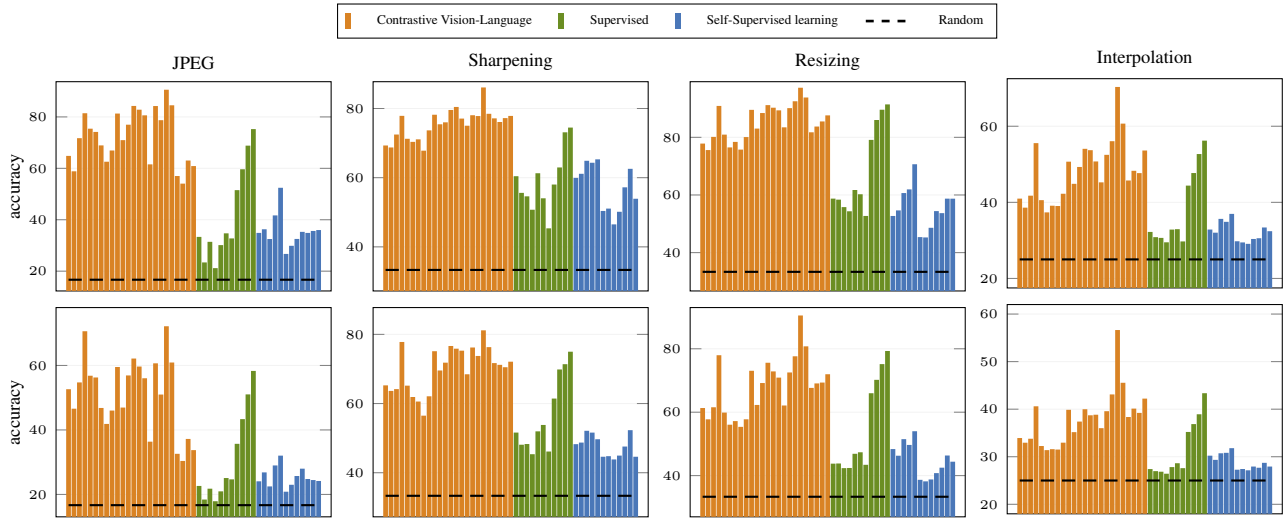}
    \vspace{-7pt}
    \caption{\textbf{Image processing-based label prediction.} Classification accuracy using a linear classifier on embeddings of different frozen visual encoders on ImageNet (top) and iNaturalist (bottom) datasets. Ordering is according to \cref{tab:models} in the supplementary material.}
    \label{fig:acc_processing_predict}
\end{figure*}

\section{Processing and acquisition traces}
\label{sec:detection}

We examine whether visual encoders retain traces of image processing and acquisition parameters. A classifier is trained to predict the metadata labels. If its performance is close to random accuracy, it indicates that the embeddings do not encode information related to processing or acquisition parameters. If the classifier performs better than random chance, it shows that information about these parameters is indeed encoded within the embeddings.

\subsection{Prediction of processing-based labels}
\label{sec:processing_prediction}

\paragraph{Setup} Given a training set $\cX_{\text{train}}$ and a test set $\cX_{\text{test}}$, we process every training image $x \in \cX_{\text{train}}$ with $p_x \sim \text{Uniform}(\cP)$, where $\text{Uniform}(\cP)$ is the uniform distribution over $\cP$, while we process all test images using $p_i \in \cP$. More specifically, we make sets $\mathcal{D}_\text{train} = \{(x, p_x)|x \in \cX_\text{train}, p_x \sim \text{Uniform}(\cP)\}$ and $\mathcal{D}_\text{test}(p_i) = \{(x, p_i)|x \in \cX_\text{test}\}$ for a given $p_i$. We then train a linear classifier to predict the metadata label. We use 20\% of the training set as the validation set to select the best learning rate and weight decay over 30 trials following~\cite{sariyildiz:hal-03929621}.\footnote{Our implementation is based on \url{https://github.com/naver/trex/tree/master/transfer}} After finding the best hyperparameters, we re-train the classifier on the whole training set.
We report the classification accuracy on the test set averaged over every possible choice of $p_i$, as well as across 10 different seeds that control the sampling of $p_x$.\footnote{Hyperparameters are tuned on the sampling of $p_x$ of the first seed.}

\paragraph{Results} \cref{fig:acc_processing_predict} shows the linear classification accuracy for predicting processing-based labels. We observe that CVL models, compared to other types of models, show a higher ability to recognize all processing-based labels; \eg CVL models are capable of exceeding $80\%$ accuracy for predicting JPEG compression, sharpening, and resizing on ImageNet. Specifically, one version of a CVL model based on ConvNeXt performs the best for all processing types. Additionally, many supervised models, mostly based on ConvNeXt, also obtain a high performance. On the other hand, self-supervised models generally perform the worst on all processing parameters.
Regarding different processing types, predicting interpolation is the hardest across all model types. However, the best performing CVL and supervised models still perform well above the random accuracy of $25\%$ on both datasets.

\input{figures/masking_with_viz_v2}

\begin{figure*}[t]
    \centering
    \vspace{-10pt}
    \input{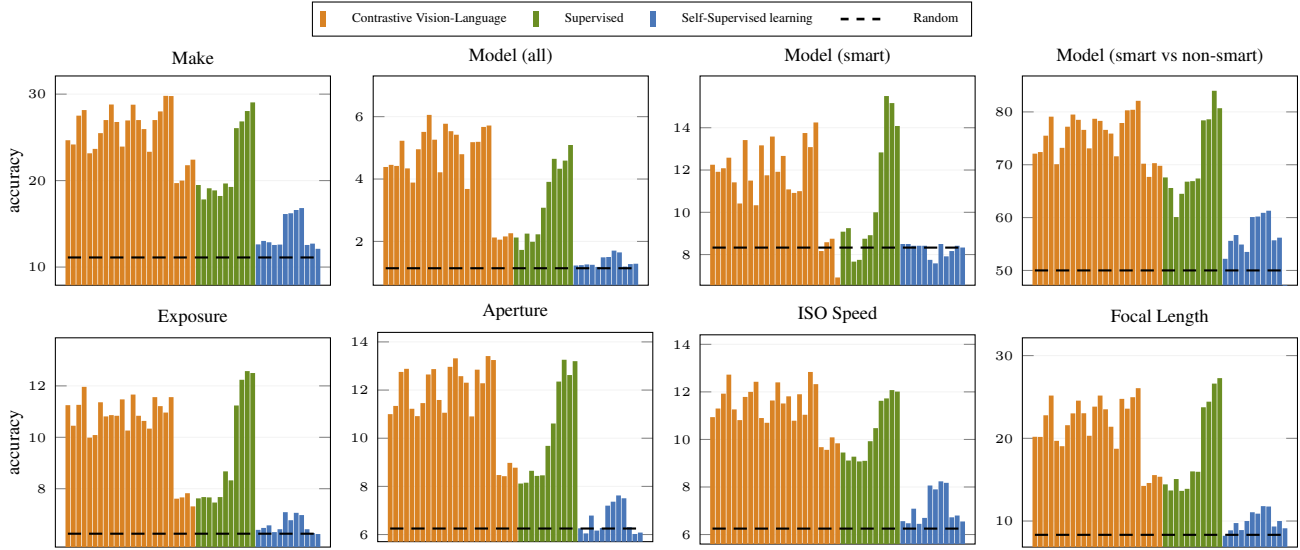}
    \vspace{-10pt}
    \caption{{\textbf{Image acquisition-based label prediction.}} Classification accuracy using a linear classifier on embeddings of frozen visual encoders with images masked at 95\% on FlickrExif dataset. Ordering is according to \cref{tab:models} in the supplementary material.}
    \vspace{-7pt}
    \label{fig:linear_probing_accuracies_ninety}
\end{figure*}

\subsection{Prediction of acquisition-based labels}
\label{sec:acquisition_prediction}

\paragraph{Setup} For each acquisition parameter $\mathcal{A}$, %
we create a training set $\cX_\text{train}$ and a test set $\cX_\text{test}$. Except for model (all) and model (smart), we only use values of $\mathcal{A}$ with at least $5,000$ images. From these, we randomly select $500$ images for the test set. The remaining images are undersampled to the size of the minority class of $\mathcal{A}$, with $200$ images per class also used for validation. For the model (all) and model (smart) parameters, due to the limited amount of data per class, we instead select classes with at least $500$ images, randomly select $500$ of these images per class, and divide these into $\cX_\text{train}$ and $\cX_\text{text}$ at a $4:1$ ratio, with $20\%$ of $\cX_\text{train}$ used for validation. 
We ensure that photographers are disjoint between the splits. We then use a linear classifier to predict the value of $\mathcal{A}$ from precomputed visual embeddings.
Similarly to \cref{sec:processing_prediction}, we select the best hyperparameters on the validation set and re-train the classifier on the whole training set.
To avoid spurious correlations between acquisition-based labels and semantic-based labels (\eg, photos of birds taken with a specific focal length), we deliberately suppress semantic information by center-masking $95\%$ of the image content.\footnote{In contrast to the version of this paper that appeared at ICCV 2025, whenever masking is applied we set all encoders' resize sizes and center crop sizes to be equal, and mask after cropping. Furthermore we now use 95\% masking, equivalent to 90\% masking in the ICCV version.} \cref{fig:masking} shows the ImageNet validation accuracy for multiple visual encoders when images are center-masked at ratios from 0 to 100\%.

\begin{table}[t]
\footnotesize
\centering
\caption{Percentage of top-k neighbors with the same model (smart vs non-smart) as the query, averaged across all queries. Results on FlickrExif.}

\setlength{\tabcolsep}{5pt}
\begin{tabular}{lllrrr}
\toprule
model & variant & class & $k=10$ & $k=25$ & $k=50$ \\
\midrule 
CLIP & ViT-L/14@336 & CVL & $70.0$ & $69.1$ & $67.6$ \\
SigLIP & ViT-B/16 & CVL & $59.0$ & $57.5$ & $56.4$ \\
\midrule
ConvNeXt & ConvNeXt-B & SUP & $57.1$ & $55.8$ & $55.2$ \\
ViT & ViT-H/14 & SUP & $53.1$ & $53.0$ & $52.4$ \\
\midrule
MoCov3 & ViT-B & SSL & $58.3$ & $56.3$ & $54.8$ \\
DINOv2 & ViT-L/14 & SSL & $54.1$ & $53.7$ & $53.0$ \\
\bottomrule
\end{tabular}

\label{tab:same_attribute_rate}
\end{table}

\begin{figure}[t]
    \centering
    \includegraphics[width=0.48\textwidth]{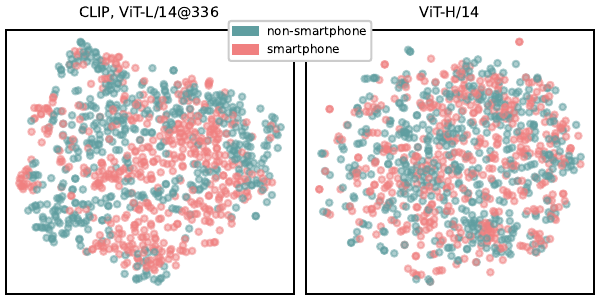}
    \vspace{-15pt}
    \caption{\textbf{t-SNE visualizations.} for two different visual encoders. Colors identify images by model (smart vs non-smart).}
    \vspace{-7pt}
    \label{fig:tsne_acquisition_attributes}
\end{figure}

\paragraph{Results}
~\cref{fig:linear_probing_accuracies_ninety} illustrates prediction accuracies for acquisition-based labels on the FlickrExif test sets.
Even with $95\%$ masking, visual encoders predict acquisition parameters well above random chance. Results are generally consistent with \cref{sec:processing_prediction} given that the highest accuracies tend to belong to CVLs and supervised ConvNeXts, albeit the best performance on some attributes belong to SSLs, and the gaps between training schemes are not as wide.
All CVLs exceed $20\%$ accuracy for make against a baseline of $11.11\%$, with many exceeding $25\%$ and some passing $30\%$. In contrast, not all SUP models can reach $20\%$, and no SSL model passes $30\%$. Furthermore, \cref{tab:same_attribute_rate} shows the rate at which an image's top-$k$ neighbors share its acquisition-based label for model (smart vs non-smart) across different encoders, without masking. The highest rates are from CVLs, which range from $59.0\%$ to $70.0\%$ compared to at most $58.3\%$ from non-CVLs. The implication is that while visual encoders are storing model information in their embeddings, this trend is strongest among CVLs. This is visualized in the t-SNE~\cite{van2008visualizing} representations in \cref{fig:tsne_acquisition_attributes}, which shows how the most and the least discriminative encoders in \cref{tab:same_attribute_rate} distribute images in their embedding spaces. While CLIP, a CVL model, shows areas of high concentration of images of similar model (smart vs non-smart), ViT, a supervised model, shows a much more even mix of points. This trend is further supported in \cref{sec:downstream_effects}.

\begin{figure*}
    \centering
    \vspace{-10pt}
    \input{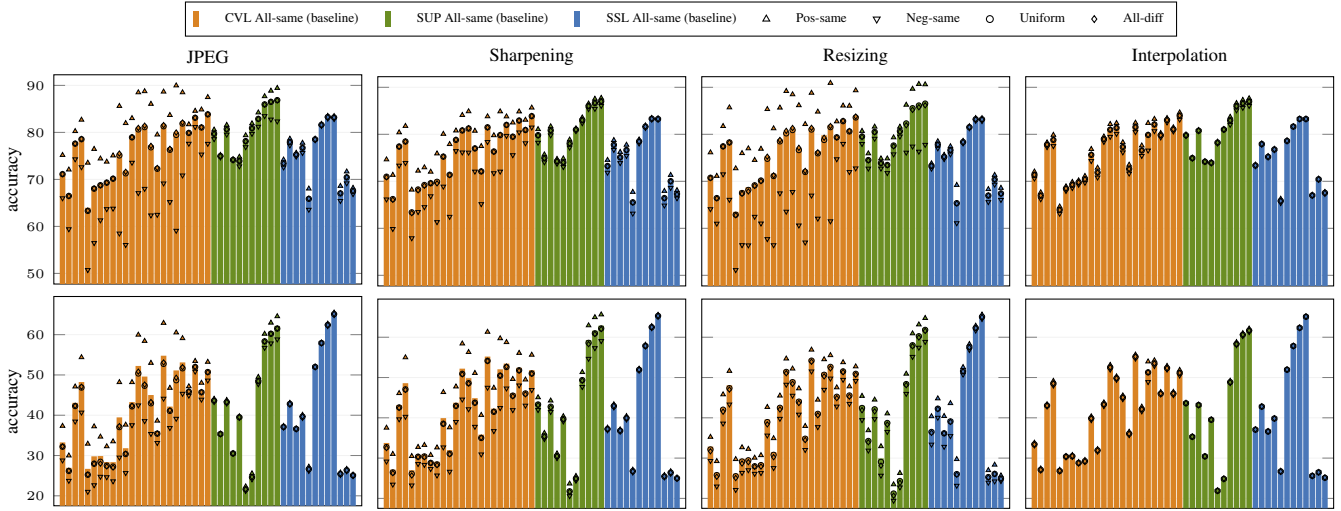}
    \vspace{-10pt}
    \caption{\textbf{Impact of image processing parameters on semantics.} Semantic label prediction accuracy on ImageNet (top) and iNaturalist (bottom) datasets in five different setups. 
    \emph{All-same (baseline)}: all training and test images share the same processing-based metadata label. 
    \emph{All-diff}: training images have the same metadata label, which is different than that of the test image.    
    \emph{Pos-same}: training images that are semantically positive to the test image have the same metadata label as the test image. 
    \emph{Neg-same}: training images that are semantically negative to the test image have the same metadata label as the test image. 
    \emph{Uniform}: the metadata labels are uniformly assigned to the training images.
    \emph{Pos-same} and \emph{neg-same} settings require an artificially created experiment where a different training set is used per test image. 
    Shown for $k=10$ and $k=1$ for the kNN classifier for ImageNet  and iNaturalist, respectively.}
    \label{fig:processing_semantic}
\end{figure*}

\section{Effect of traces on downstream tasks}

\label{sec:downstream_effects}

While we have shown that metadata labels can be predicted from image embeddings, in this section, we investigate whether this information can impact downstream applications that rely on the ability to predict semantic labels. We focus on two main tasks: image classification and near-duplicate image retrieval.

\subsection{Processing distracts semantic predictions}

\label{sec:processing_semantic}
To analyze the impact of processing parameters on semantic predictions, we use a kNN classifier to predict the semantic label of each image. In that way, we assess the representation capability of the models to bring closer images with the same semantic label.

\vspace{-10pt}
\paragraph{Setup} Given a training set $\cX_{\text{train}}$ and a test set $\cX_{\text{test}}$, the same as in \cref{sec:processing_prediction}, we measure the classification accuracy of predicting the semantic label. We perform this in five setups, where \emph{a separate training set is formed per test image}.

\begin{enumerate}[label=(\roman*)]
    \item \emph{All-same (baseline):} processing all training $\cX_{\text{train}}$ and test $\cX_{\text{test}}$ images with an identical processing value $p_i$. 
    It serves as a baseline for reference, where classification performance is fully determined by the semantic similarity between image embeddings.     
   \item \emph{All-diff:} processing all training $\cX_{\text{train}}$ images with processing value $p_j$ and test $\cX_{\text{test}}$ images with processing value $p_i$, for $i\neq j$. 
    It reflects the common setup of existing robustness studies~\cite{hendrycks2019benchmarking}.    
    \item \emph{Pos-same:} processing a test image $x \in \cX_\text{test}$ and training images that share the same semantic label (positive) as $x$ with $p_i$, while processing training images that have a different semantic label (negative) with processing value $p_j$ with $i \ne j$. 
    It investigates whether semantic positives (negatives) being metadata positives (negatives) affects the accuracy, compared to the baseline.
    \item \emph{Neg-same:} processing a test image $x \in \cX_\text{test}$ and training images that do not have the same semantic label (negative) as $x$ with $p_i$, while processing training images that have the same semantic label (positive) as $x$ with processing value $p_j$ with $i \ne j$.
    It investigates whether semantic positives (negatives) being metadata negatives (positives) affects the accuracy, compared to the baseline.
    \item \emph{Uniform:} similarly to \cref{sec:processing_prediction}, processing $x \in \cX_{\text{test}}$ with $p_i$ while $x' \in \cX_{\text{train}}$ is processed with $p_{x'} \sim \text{Uniform}(\cP)$. 
    This reflects a more realistic scenario.
\end{enumerate}
\noindent We report classification accuracy averaged across all possible combinations of $p_i$ and $p_j$ for each setup, as well as for $10$ random seeds for uniform sampling of $p_{x'}$.

\begin{figure*}[t]
    \centering
    \vspace{-5pt}
    \resizebox{\linewidth}{!}{
      \newcommand{\imagebox}[3][]{%
    \begin{tikzpicture}[baseline=(img.center)]
        \node[inner sep=0pt, draw, line width=2mm, draw=#3, name=instance] (img) {
            \adjustbox{width=0.075\linewidth, height=0.075\linewidth}{
                \includegraphics{#2}
            }
        };
        \ifx\\#1\\
        \else
            \node[anchor=west, text=black, font=\scriptsize, yshift=0.93cm, xshift=-0.9cm, align=left] at (img) {\textbf{prec@10 = #1\%}};
        \fi
    \end{tikzpicture}
}

\newcommand{\jpgbox}[1]{%
    \begin{tikzpicture}[baseline=(img.center)]
        \node[inner sep=0pt, name=instance] (img) {
            \adjustbox{height=0.05\linewidth}{
                \includegraphics{#1}
            }
        };
    \end{tikzpicture}
}

\hspace{-15pt}
\setlength{\tabcolsep}{0pt}
\renewcommand{\arraystretch}{1.2} 
\begin{tabular}{ccc@{\hspace{0.5cm}} c@{\hspace{0.1cm}} *{10}{c} c} 
    \tikzsetnextfilename{visualization_effects_1}
    \jpgbox{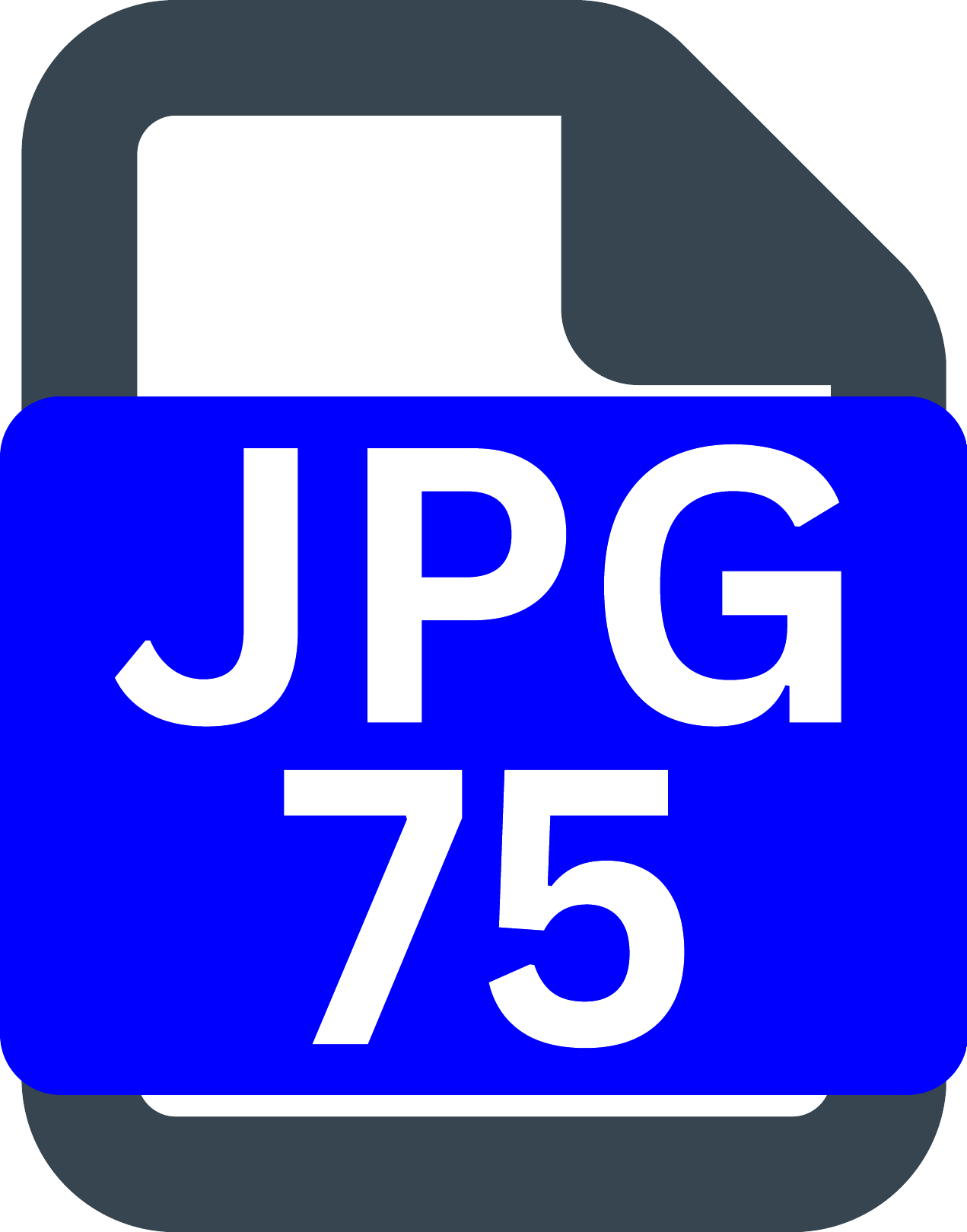} &
    \tikzsetnextfilename{visualization_effects_2}
    \jpgbox{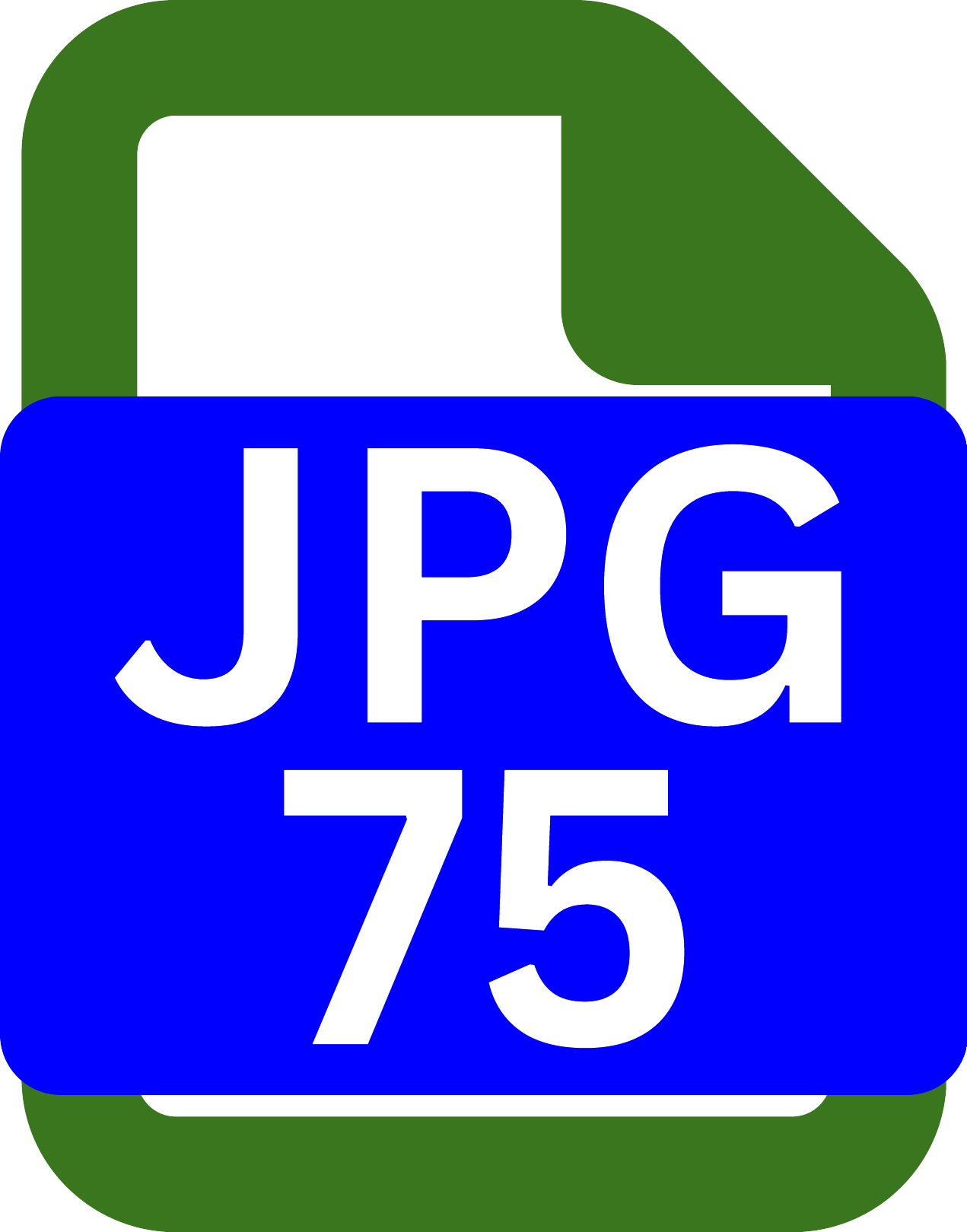} &
    \tikzsetnextfilename{visualization_effects_3}
    \jpgbox{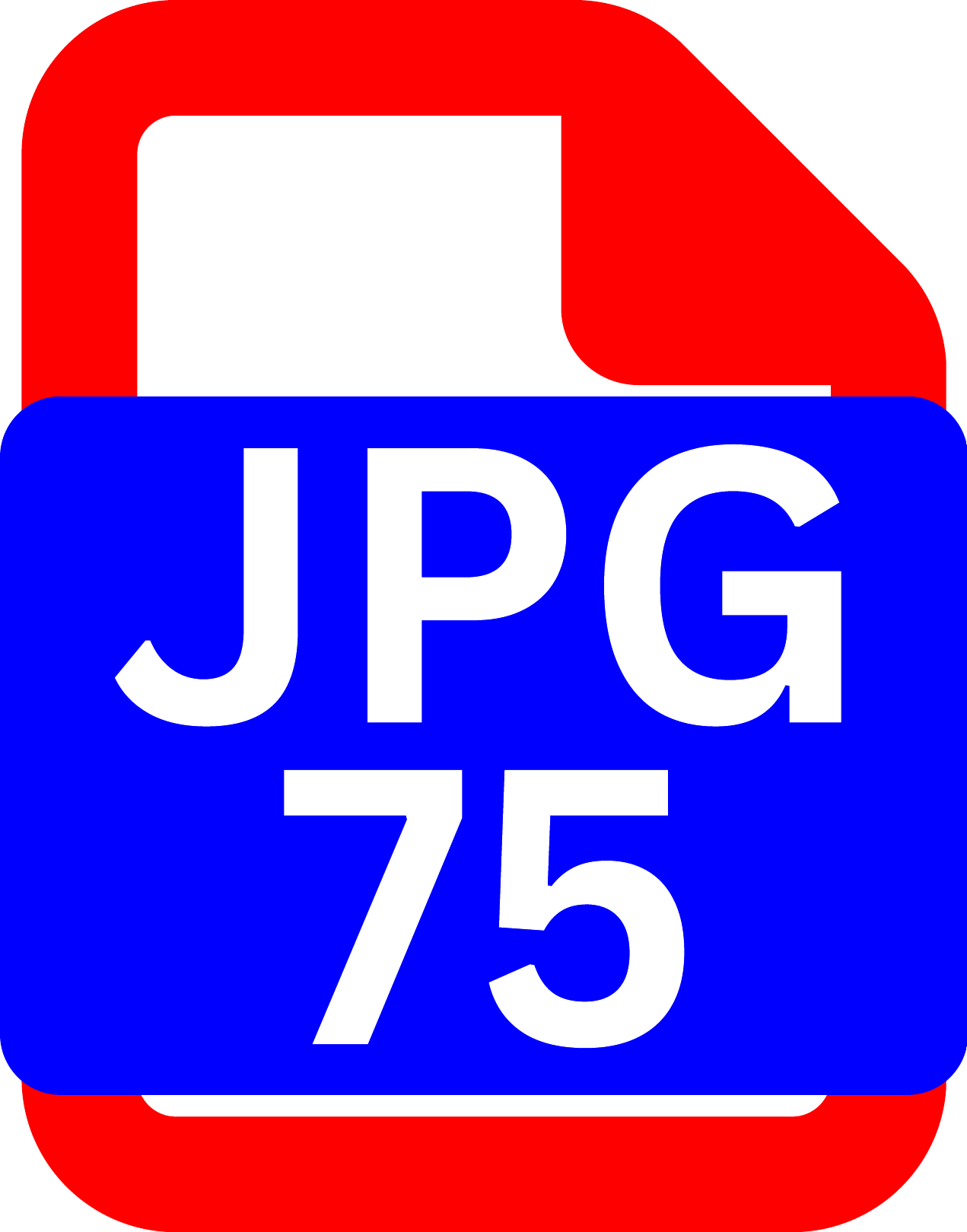} &
    \tikzsetnextfilename{visualization_effects_4}
    \imagebox[40]{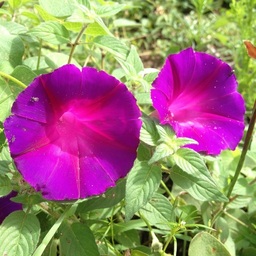}{lightblue} &
    \tikzsetnextfilename{visualization_effects_5}
    \imagebox[]{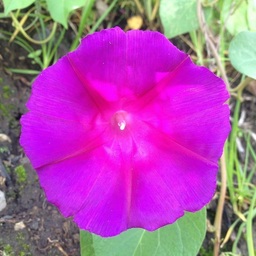}{lightgreen} &
    \tikzsetnextfilename{visualization_effects_6}
    \imagebox[]{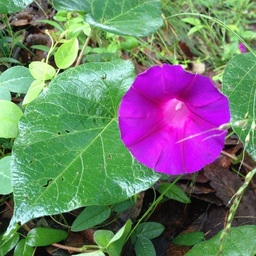}{lightgreen} &
    \tikzsetnextfilename{visualization_effects_7}
    \imagebox[]{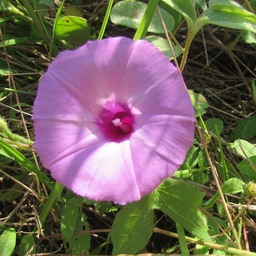}{lightred} &
    \tikzsetnextfilename{visualization_effects_8}
    \imagebox[]{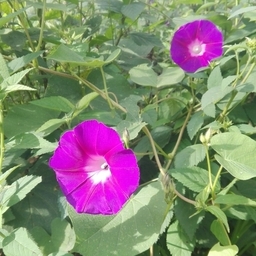}{lightgreen} &
    \tikzsetnextfilename{visualization_effects_9}
    \imagebox[]{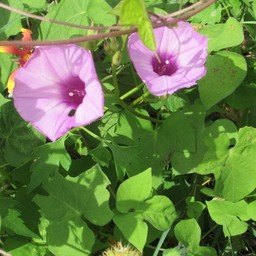}{lightred} &
    \tikzsetnextfilename{visualization_effects_10}
    \imagebox[]{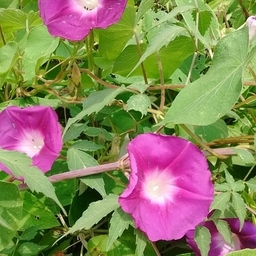}{lightgreen} &
    \tikzsetnextfilename{visualization_effects_11}
    \imagebox[]{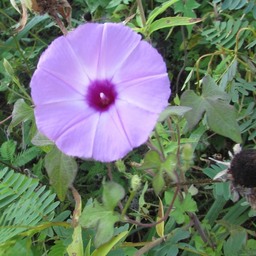}{lightred} &
    \tikzsetnextfilename{visualization_effects_12}
    \imagebox[]{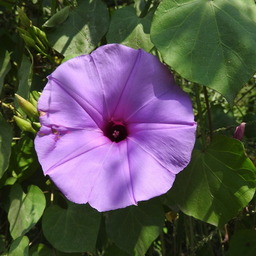}{lightred} &
    \tikzsetnextfilename{visualization_effects_13}
    \imagebox[]{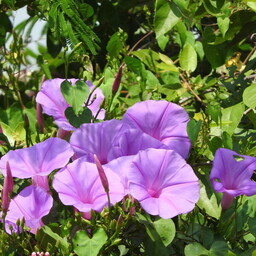}{lightred} &
    \tikzsetnextfilename{visualization_effects_14}
    \imagebox[]{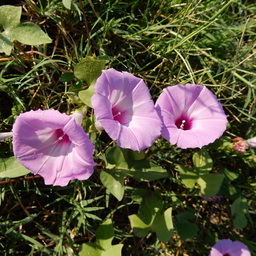}{lightred} 

    \\

    \tikzsetnextfilename{visualization_effects_15}
    \jpgbox{images/processing_teaser/q_75.pdf} &
    \tikzsetnextfilename{visualization_effects_16}
    \jpgbox{images/processing_teaser/p_75.pdf} &
    \tikzsetnextfilename{visualization_effects_17}
    \jpgbox{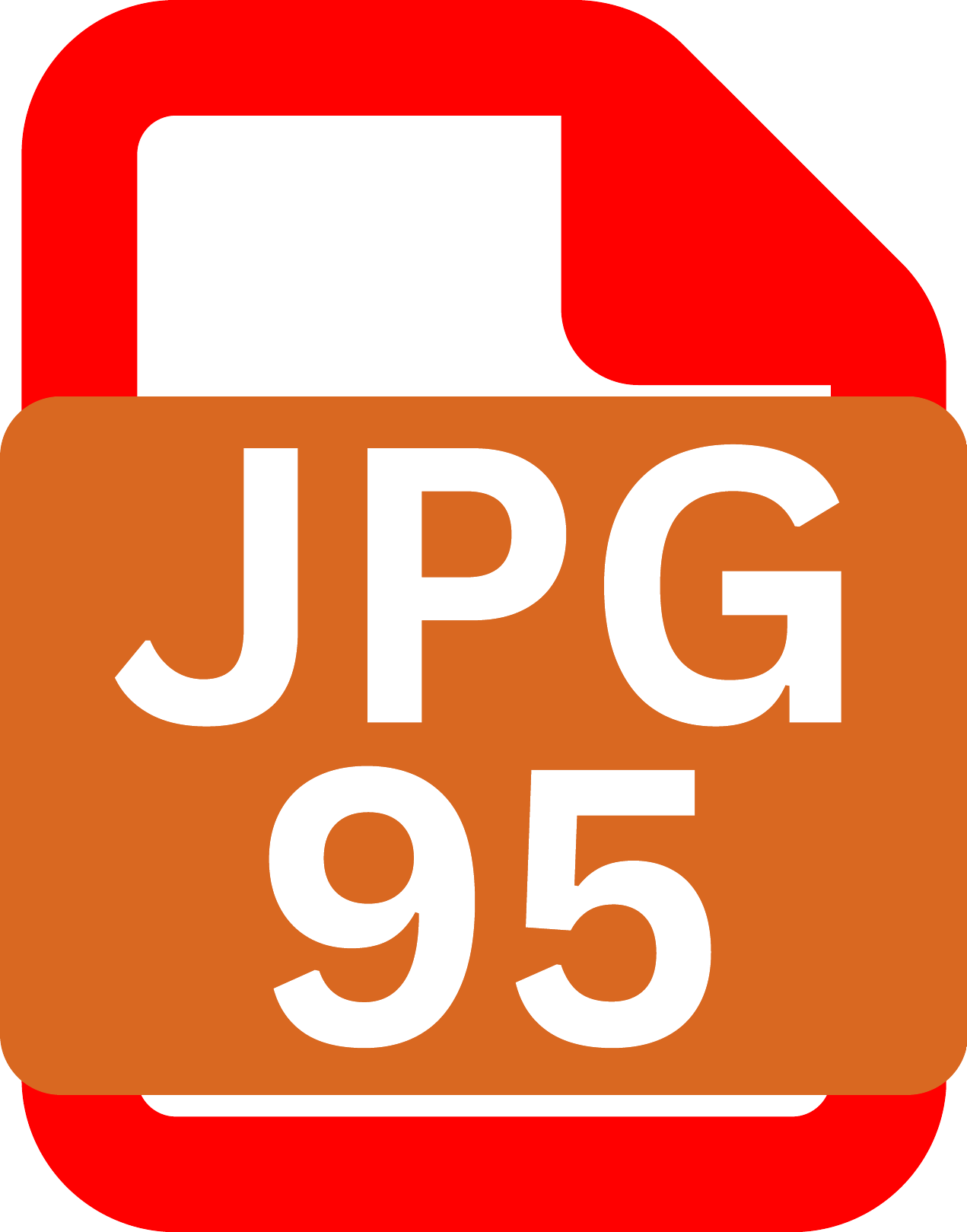} &
    \tikzsetnextfilename{visualization_effects_18}
    \imagebox[80]{images/processing_teaser/1000/query/query.jpg}{lightblue} &
    \tikzsetnextfilename{visualization_effects_19}
    \imagebox[]{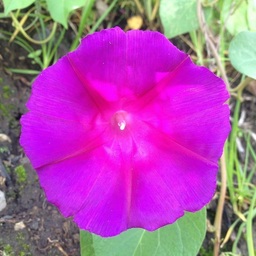}{lightgreen} &
    \tikzsetnextfilename{visualization_effects_20}
    \imagebox[]{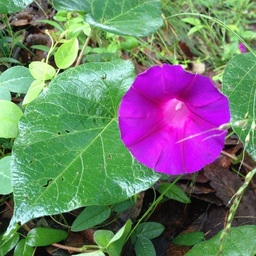}{lightgreen} &
    \tikzsetnextfilename{visualization_effects_21}
    \imagebox[]{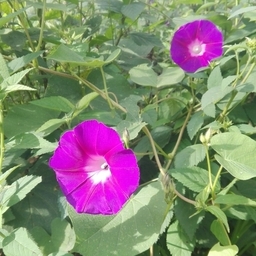}{lightgreen} &
    \tikzsetnextfilename{visualization_effects_22}
    \imagebox[]{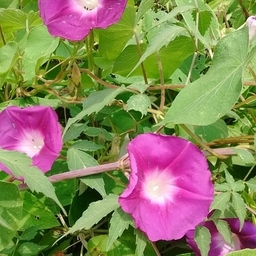}{lightgreen} &
    \tikzsetnextfilename{visualization_effects_23}
    \imagebox[]{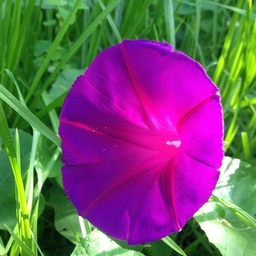}{lightgreen} &
    \tikzsetnextfilename{visualization_effects_24}
    \imagebox[]{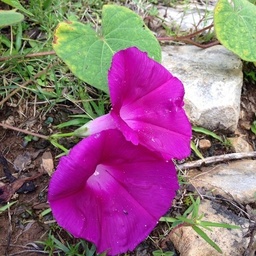}{lightgreen} &
    \tikzsetnextfilename{visualization_effects_25}
    \imagebox[]{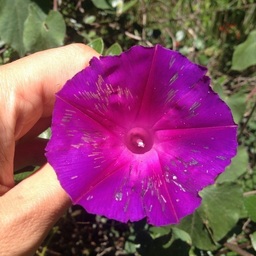}{lightgreen} &
    \tikzsetnextfilename{visualization_effects_26}
    \imagebox[]{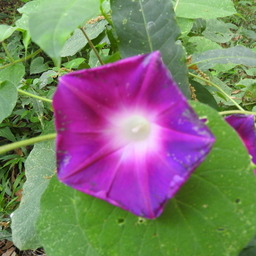}{lightgreen} &
    \tikzsetnextfilename{visualization_effects_27}
    \imagebox[]{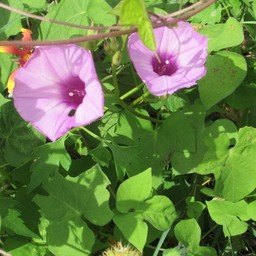}{lightred} &
    \tikzsetnextfilename{visualization_effects_28}
    \imagebox[]{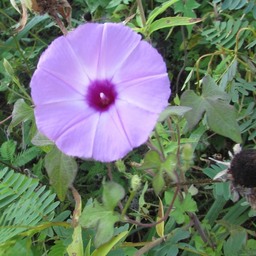}{lightred} 
    
    \\

    \tikzsetnextfilename{visualization_effects_29}
    \jpgbox{images/processing_teaser/q_75.pdf} &
    \tikzsetnextfilename{visualization_effects_30}
    \jpgbox{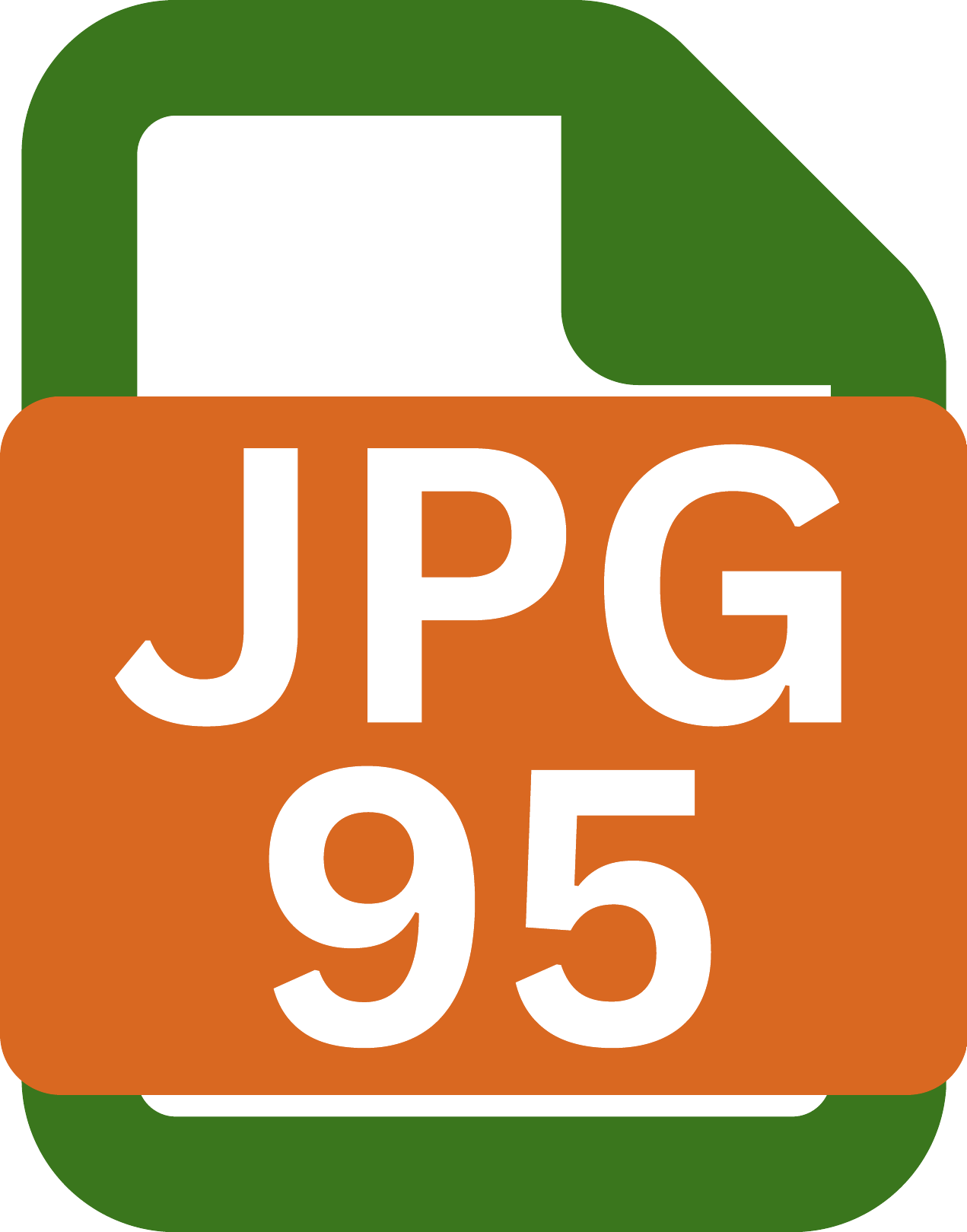} &
    \tikzsetnextfilename{visualization_effects_31}
    \jpgbox{images/processing_teaser/n_75.pdf} &
    \tikzsetnextfilename{visualization_effects_32}
    \imagebox[20]{images/processing_teaser/1000/query/query.jpg}{lightblue} &
    \tikzsetnextfilename{visualization_effects_33}
    \imagebox[]{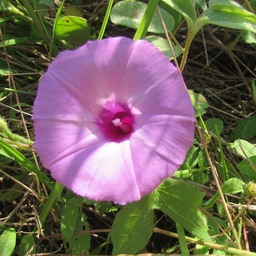}{lightred} &
    \tikzsetnextfilename{visualization_effects_34}
    \imagebox[]{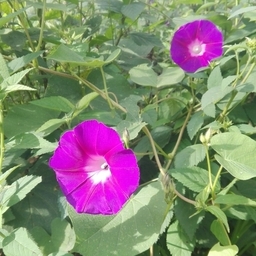}{lightgreen} &
    \tikzsetnextfilename{visualization_effects_35}
    \imagebox[]{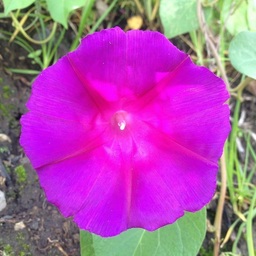}{lightgreen} &
    \tikzsetnextfilename{visualization_effects_36}
    \imagebox[]{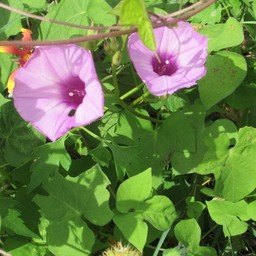}{lightred} &
    \tikzsetnextfilename{visualization_effects_37}
    \imagebox[]{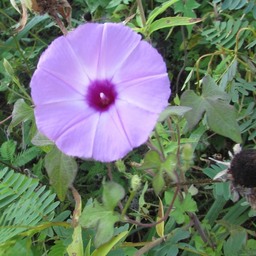}{lightred} &
    \tikzsetnextfilename{visualization_effects_38}
    \imagebox[]{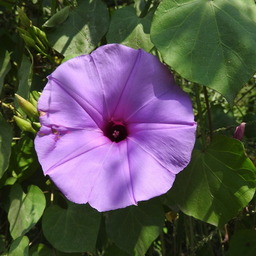}{lightred} &
    \tikzsetnextfilename{visualization_effects_39}
    \imagebox[]{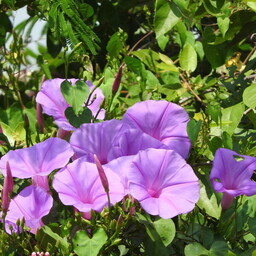}{lightred} &
    \tikzsetnextfilename{visualization_effects_40}
    \imagebox[]{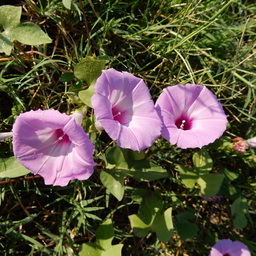}{lightred} &
    \tikzsetnextfilename{visualization_effects_41}
    \imagebox[]{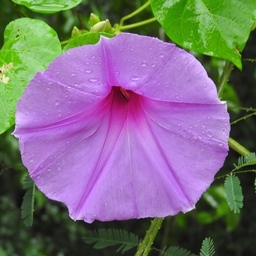}{lightred} &
    \tikzsetnextfilename{visualization_effects_42}
    \imagebox[]{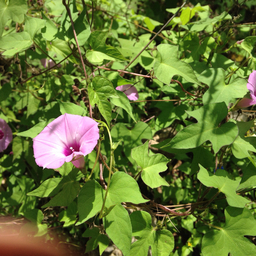}{lightred} 
\end{tabular}

    }
    \caption{\textbf{Visualization of top-10 nearest neighbors of a query.} in three different setups: \emph{All-same (baseline)} (top), \emph{Pos-same} (middle), \emph{Neg-same} (bottom) using CVL ConvNeXt-L model. 
    Green/red borders indicate a positive/negative image to the query  according to the semantic label, while the icons on the left indicate the metadata label (JPEG compression) for the \textbf{{\color{lightblue}query}}, \textbf{{\color{lightgreen}positive}}, and \textbf{{\color{lightred}negative}} images.
    \label{fig:visualization_effects}
    }
\end{figure*}

\vspace{-10pt}
\paragraph{Results} \cref{fig:processing_semantic} illustrates the impact of metadata labels related to processing parameters on semantic recognition.
When semantic and metadata labels are coupled and change together (\emph{pos-same}, \emph{neg-same}), performance is significantly affected for certain processing types and models.
Although these setups are extreme, they represent corner cases that could still arise in real-world scenarios for a specific test image.
JPEG compression has the greatest impact, followed by sharpening and resizing. Interpolation mostly has an insignificant effect on semantics. The greatest sensitivity is observed in CVLs, followed by some supervised models, mainly the ones based on the ConvNeXt architecture. Self-supervised models are the least affected, providing hints on heavy geometric and photometric training augmentations causing the different behavior.
\looseness=-1 The \emph{all-diff} and \emph{uniform} setups perform similarly, typically slightly below the baseline. This suggests that the perspective presented in this study differs from the traditional robustness assessments in prior research.

\looseness=-1 We visualize the top-10 nearest neighbors of a query image in three different setups in \cref{fig:visualization_effects} for one of the most impacted models, \ie the CVL ConvNeXt-L model. The traces of metadata labels in the representation space significantly affect the quality of the nearest images. 
\looseness=-1 Evidently, distance in the representation space is a function of both semantic and metadata labels for some foundation models.

\subsection{Camera model distracts semantic retrieval}
Using the PairsCams dataset, we investigate the impact of image acquisition parameters, in particular whether the camera was a modern smartphone, on the task of near-duplicate image retrieval.

\vspace{-10pt}
\paragraph{Setup}

Given a query image of an object or scene captured by one camera type, the objective is to retrieve the corresponding image of the same object or scene (positive) taken by a different camera type. We consider two different settings for the negative images in the retrieval database: (i) \emph{same} -- negatives are captured by the same camera type as the query, or (ii) \emph{different} -- negatives are captured by a different camera type. We evaluate performance based on the average $\text{recall}@1$ under both settings.

\vspace{-10pt}
\paragraph{Results}

\begin{figure}[t]
    \centering
    \input{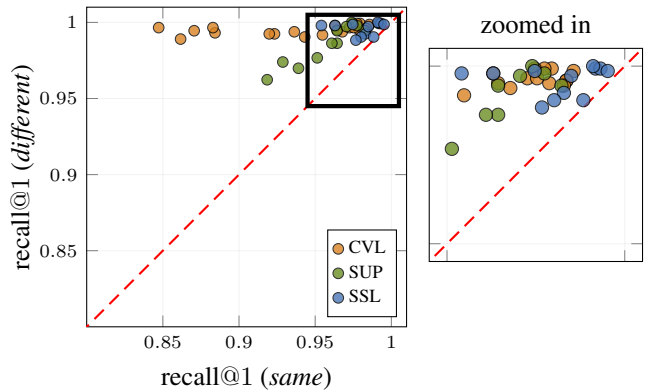}
\pgfplotsset{every tick label/.append style={font=\footnotesize}}
\pgfplotsset{select coords between index/.style 2 args={
    x filter/.code={
        \ifnum\coordindex<#1\def\pgfmathresult{}\fi
        \ifnum\coordindex>#2\def\pgfmathresult{}\fi
    }
}}
\pgfplotsset{minor grid style={solid,gray,opacity=0.1}}
\pgfplotsset{major grid style={solid,gray,opacity=0.1}}

\pgfdeclarelayer{foreground}
\pgfsetlayers{main,foreground}

\begin{tabular}{c@{\nxssp}c@{\nxssp}}

\hspace{-15pt}
\tikzsetnextfilename{ndidscatter_1}
\begin{tikzpicture}[baseline={(current bounding box.north)}]
    \begin{axis}[
        width=.7\linewidth,
        height=.7\linewidth,
        ylabel={$\text{recall}@1$ (\emph{different})},
        xlabel={$\text{recall}@1$ (\emph{same})},
        ymin=0.8,
        ymax=1.01,
        xmin=0.8,
        xmax=1.01,
        grid=both,
        tick label style={font=\scriptsize},
        ytick={0.85, 0.9, 0.95, 1.},
        xtick={0.85, 0.9, 0.95, 1.},
        ylabel near ticks, xlabel near ticks, 
        ylabel style={yshift=-5pt},
        legend entries={CVL, SUP, SSL},
        legend pos=south east,
        legend style={cells={anchor=west}, font=\scriptsize, fill opacity=1.0, row sep=-1pt, inner sep=2pt},
        grid=both,
        grid style={color=lightgray!60, dash pattern=on 2pt off 2pt},
        ]
        
        \addplot[only marks, mark options={draw=black}, color=vlmcolor, solid, mark=*, opacity=0.8, mark size=2] table[y=diff, x expr={\thisrow{id} / (\thisrow{show} == 0)}] \ndidscatternew;
        
        \addplot[only marks, mark options={draw=black}, color=supervisedcolor, opacity=0.8, solid, mark=*, mark size=2] table[y=diff, x expr={\thisrow{id} / (\thisrow{show} == 1)}] \ndidscatternew;
        
        \addplot[only marks, mark options={draw=black}, color=sslcolor, solid, mark=*, opacity=0.8, mark size=2] table[y=diff, x expr={\thisrow{id} / (\thisrow{show} == 2)}] \ndidscatternew; 
        
        \addplot[red, thick, mark=none, dash pattern=on 5pt off 3pt] coordinates {(-0, -0) (1.01, 1.01)};
        
        \pgfonlayer{foreground}
        \draw[black, ultra thick] (axis cs:0.945,0.945) rectangle (axis cs:1.005,1.005);
        \endpgfonlayer
        
    \end{axis}
\end{tikzpicture}

&

\hspace{-5pt}
\tikzsetnextfilename{ndidscatter_2}
\begin{tikzpicture}[baseline={(current bounding box.north)}]
    \begin{axis}[
        width=.53\linewidth,
        height=.53\linewidth,
        ymin=0.945,
        ymax=1.005,
        xmin=0.945,
        xmax=1.005,
        grid=both,
        tick label style={font=\scriptsize},
        title={zoomed in},
        title style={yshift=-4pt},
        ytick={0.85, 0.9, 0.95, 1.},
        xtick={0.85, 0.9, 0.95, 1.},
        yticklabel=\empty,
        xticklabel=\empty,
        ylabel near ticks, xlabel near ticks, 
        grid=both,
        grid style={color=lightgray!60, dash pattern=on 2pt off 2pt},
        ]

        \addplot[only marks, mark options={draw=black}, color=vlmcolor, solid, mark=*, opacity=0.8, mark size=2.5] table[y=diff, x expr={\thisrow{id} / (\thisrow{show} == 0)}] \ndidscatternew;
        
        \addplot[only marks, mark options={draw=black}, color=supervisedcolor, solid, mark=*, opacity=0.8, mark size=2.5] table[y=diff, x expr={\thisrow{id} / (\thisrow{show} == 1)}] \ndidscatternew;
        
        \addplot[only marks, mark options={draw=black}, color=sslcolor, solid, mark=*, opacity=0.8, mark size=2.5] table[y=diff, x expr={\thisrow{id} / (\thisrow{show} == 2)}] \ndidscatternew; 
        
        \addplot[red, thick, mark=none, dash pattern=on 5pt off 3pt] coordinates {(-0, -0) (1.01, 1.01)};

    \end{axis}
\end{tikzpicture}
\end{tabular}
    \vspace{-10pt}
    \caption{\textbf{Impact of image acquisition parameters on semantics.} Retrieval performance measured by $\text{recall}@1$ using the PairsCams dataset for the settings where the negatives are captured either by the \emph{same} or a \emph{different} camera type than the query.
    Each point corresponds to a different visual encoder. 
    The diagonal line is provided as a reference to indicate a lack of influence from the metadata label, \ie model (smart vs non-smart).}
    \label{fig:scatterplot}
    \vspace{-5pt}
\end{figure}

Results are presented in \cref{fig:scatterplot}. 
All points are above the diagonal, suggesting the influence of the metadata label on the retrieval performance.
Vision encoders consistently fail to retrieve the image of the same object or scene in the presence of negatives taken with the same camera type.
The experiment reveals that CVLs have the highest disparity between the two settings. 
A model with near-perfect $\text{recall}@1$ can see a performance drop below $0.85$ simply by having the negatives share the same camera type as the queries.
See \cref{fig:fig1} for a qualitative example; the near-identical positive is not retrieved first due to the metadata label mismatch.

\section{Related work}
\label{sec:rel_work}

\paragraph{Robustness benchmarks and mitigation}
There has been much effort on the improvement of visual encoders' robustness under shifts from their training distributions. This includes the design of benchmarks that evaluate robustness on hard conditions~\cite{hendrycks2019benchmarking,hendrycks2021natural,hendrycks2021many,geirhos2018imagenet,baek2024unexplored,li2023imagenet}, with some specializing on corruptions and perturbations~\cite{hendrycks2019benchmarking}, stylization techniques~\cite{geirhos2018imagenet}, JPEG compressions~\cite{ghosh2018robustness}, adversarial and out-of-distribution examples~\cite{hendrycks2021natural}, attribute editing~\cite{li2023imagenet}, and environment and sensor shifts~\cite{baek2024unexplored}. Other work focuses specifically on CVL models under similar conditions \ie adversarial examples~\cite{mao2022understanding}, perturbations~\cite{schiappa2022robustness}, JPEG compression~\cite{chen2023understanding}, and multiple aspects of robustness simultaneously ~\cite{liang2021multibench}. Additionally, several solutions for improving robustness, involving the use of strong augmentations~\cite{geirhos2018imagenet,hendrycks2019benchmarking}, a more elaborate training process~\cite{zheng2016improving}, learning artifact correction~\cite{ehrlich2021analyzing}, test-time adaptation~\cite{schneider2020improving} and training~\cite{wang2020tent,osowiechi2025watt}, or prompting~\cite{li2024one} and adversarial fine-tuning~\cite{schlarmann2024robust}, have been shown to improve CVL zero-shot performance. Unlike prior work, we analyze images' processing and acquisition attributes and the ability of pretrained models to capture such information. We show that such signals can dominate representations and distract semantics, even though they are imperceptible to the naked eye. Also, we do not consider training images derived from a ``clean'' in-domain distribution since all images carry their unique processing and acquisition attributes, with only a small fraction being controllable (\eg, JPEG compression).

\vspace{-5pt}
\paragraph{Analysis of visual encoder robustness} 
Analyzing the behavior of visual encoders, Goodfellow~\etal~\cite{goodfellow2014explaining}  investigate the impact of inserting human-imperceptible noise into test images during evaluation. It is found that by altering just one pixel, model predictions can be changed~\cite{su2019one}. Other analysis involve texture-shape bias~\cite{geirhos2018imagenet}, translation/shift invariance and positional encoding~\cite{kayhan2020translation,islam2020much,zhang2019making}, comparison between architecture families~\cite{naseer2021intriguing}, or emergent properties from large-scale pretraining~\cite{shtedritski2023does}. Closely related to our work is model performance assessment under different JPEG compression~\cite{ehrlich2021analyzing,chen2023understanding}, while recent work~\cite{grommelt2024fake} highlights that discrepancies in JPEG compression can affect performance evaluation, resulting in misleading outcomes.

\vspace{-5pt}
\paragraph{Attribute detection} Similar to our experiments on metadata label prediction, previous work has explored attribute prediction tasks, such as camera model identification~\cite{bayar2018towards,manisha2022beyond,rafi2021remnet} 
or camera parameters prediction 
(\eg focal length and radial distortion~\cite{veicht2024geocalib} or camera tilt and roll~\cite{lopez2019deep}). There is also work on aligning Exif tags with visual content via CVLs~\cite{zheng2023exif}, as well as predicting image compression~\cite{dumas2019context} and coding~\cite{li2018fully}. %
While our experiments share clear similarities,  this typically involves training models with supervised labels to predict image attributes. In contrast, we reveal that this type of information is already contained in the visual representations, which can predict the processing and acquisition attributes without any further training or fine-tuning.

\vspace{-5pt}
\paragraph{Visual bias}
Prior work~\cite{torralba2011unbiased,liu2024decade} has shown that dataset-specific biases allow a classifier to predict an image's dataset of origin well above random. 
To combat such kinds of biases, several methods for debiasing at training time have been proposed, like leveraging labels of attributes introducing bias~\cite{kim2019learning,sagawa2020distributionally,wang2020towards,hong2021unbiased,barbano2022unbiased} or pseudo-labels derived from the biased classifiers~\cite{bahng2020learning,nam2020learning,cadene2019rubi,sarridis2024flac}. More recent approaches employ external models to discover and then suppress biases by extracting common textual keywords from misclassified examples~\cite{kim2024discovering}, or directly using large language models for bias discovery~\cite{d2024openbias,sarridis2025mavias,guimard2025classifier}.  Our evaluation setup for analyzing the impact of metadata labels on semantic prediction shares similarities with approaches in this field~\cite{bahng2020learning,sagawa2020distributionally,hong2021unbiased}. %

\section{Discussion and conclusion}

\paragraph{Conclusions} 
We have shown that metadata labels, such as acquisition and processing parameters, are encoded in the representations of foundational visual encoders, especially those trained with CVL loss. Our analysis revealed that  traces of such labels can impact semantic recognition performance, depending on the relative distribution between semantic and metadata labels. This effect, however, is not uniform across all models: it is most pronounced in CVLs, while SSL encoders exhibit it to a lesser extent. This discrepancy may be attributed to the drastic pixel-wise and geometric augmentations in SSL, which CVLs do not employ. To verify this assumption, %
in~\cref{sec:augmentation_impact}, we train from scratch and finetune a CVL model with and without augmentations. We show that the introduction of augmentations reduces the metadata information encoded in the representations.

\vspace{-7pt}
\paragraph{Limitations}
While we have identified that metadata labels are encoded in foundational visual encoders and provided hints about potential causes, we cannot definitively pinpoint the source of the problem. Investigating this further is challenging due to the cost of retraining such models and the frequent use of private datasets and undisclosed implementation details. Although we do not propose specific mitigation techniques, we highlight the issue as an important area for future research.

\vspace{-7pt}
\paragraph{Negative implications}  Metadata labels leaving traces in visual encoders to the point of overshadowing semantic information  can lead to unpredictable outcomes, compromising generalizability, robustness, and potentially undermining the trustworthiness of the models. More critically, this effect could be exploited maliciously; for instance, an adversarial attack may manipulate metadata to intentionally mislead or deceive a model, posing risks in sensitive domains like healthcare, surveillance, or autonomous systems.

\vspace{-7pt}
\paragraph{Positive implications} There are some applications that can benefit from the implications revealed in this work. For example, recent methods for synthetic (deepfake) image detection leverage frozen CVL models for detection~\cite{ojha2023towards,sha2023fake} and localization~\cite{smeu2024declip}. Interestingly, CVL models can be used as a kNN classifier, \ie without training a linear classifier~\cite{ojha2023towards}. We presume that the capability of CVL models to detect deepfakes shares the same underlying root enabling attribute-based detection. %

\section*{Acknowledgments}
This work was supported by the Junior Star GACR GM 21-28830M, the Horizon MSCA-PF grant No. 101154126, the Czech Technical University in Prague grant No. SGS23/173/OHK3/3T/13, JSPS KAKENHI No. JP23H00497 and JP22K12091, JST CREST Grant No.~JPMJCR20D3, and JST FOREST Grant No.~JPMJFR216O. We acknowledge VSB – Technical University of Ostrava, IT4Innovations National Supercomputing Center, Czech Republic, for awarding this project access to the LUMI supercomputer, owned by the EuroHPC Joint Undertaking, hosted by CSC (Finland) and the LUMI consortium through the Ministry of Education, Youth and Sports of the Czech Republic through the e-INFRA CZ (grant ID: 90254).

We thank Jan Franců, Yannis Kalantidis, Symeon Papadopoulos, and Vladimir Risojević for many helpful comments.

{
    \small
    \bibliographystyle{ieeenat_fullname}
    \bibliography{main}
}

\clearpage
\appendix
\clearpage
\appendix
\setcounter{page}{1}
\setcounter{section}{0}
\setcounter{table}{0}
\setcounter{figure}{0}
\renewcommand{\thefigure}{\Alph{figure}}
\renewcommand{\thetable}{\Alph{table}}
\maketitlesupplementary

\section{Impact of augmentations on encoding of metadata in CVLs}
\label{sec:augmentation_impact}
To verify our assumption that one of the main reasons that SSL models generally encode less metadata information is the use of heavy augmentations, we train a CVL model using OpenCLIP~\cite{cherti2023reproducible} both without heavy augmentation (default OpenCLIP setup) and with DINOv2~\cite{oquab2024dinov2} style color augmentations.\footnote{We use color jitter, random grayscaling, and random blurring.} We train a ViT-B/32 CVL model both from scratch and finetuned from the LAION-2B checkpoint on the YFCC15M dataset~\cite{radford2021learning}. We follow the default training hyperparameters from OpenCLIP, which is training for $32$ epochs with a batch size of $32k$ using the AdamW~\cite{loshchilov19adamw} optimizer. We utilize the cosine scheduler with a warmup of $2,000$ iterations. Learning rate is set to $5e^{-4}$ and weight decay is equal to $0.2$.

\cref{fig:augmentations_processing_prediction} and~\cref{fig:linear_probing_accuracies_aug} present the results for processing-based and acquisition-based metadata label prediction, respectively. Augmentations greatly reduce the accuracy of processing-based metadata label prediction. We observe a similar trend for acquisition-based metadata labels, though not to the same extent and consistency.

\looseness=-1~\cref{fig:augmentations_processing_influence} presents the results for the effects of processing on the downstream tasks. Augmentations greatly reduce the influence of processing parameters on the prediction of semantics.

Although these results point to the lack of heavy augmentations in CVLs as one of the main reasons for their strong encoding of metadata information, further investigation is still necessary. This is because some CVL models, like SigLIP~\cite{zhai2023sigmoid}, encode less metadata information on attributes like JPEG, although they do not employ heavy augmentations.

\section{Parameters description}

\label{supp:parameters_description}

We consider the following processing and acquisition parameters.

\begin{itemize}    
\item \textbf{JPEG compression} is one of the most common operations that will be applied to an image after its acquisition. JPEG applies lossy compression where the amount of compression can be controlled by the quality and chroma-subsampling parameters. To investigate the influence of JPEG compression on image representations, we recompress images using $\text{quality} \in \{75,85,95\}$ and $\text{chroma-subsampling} \in \{\text{4:2:0}, \text{4:4:4}\}$, which gives $|\cP| = 6$ possible processing parameter values.

\item \textbf{Sharpening} corrects pixel values such that the image appears sharper, and is commonly automatically applied by different services~\cite{flickrblog}. We use unsharp mask based sharpening of an image $\cI$ given as $\text{sharp}(\cI) = \alpha\cI + (1 - \alpha)\text{blur}(\cI)$ where $\alpha$ controls the sharpness of the image. $\alpha=1$ gives the original image, while $\alpha > 1$ gives a sharper image. For $\cP$ we consider a set of processing parameter values given when $\alpha \in \{1, 2, 4\}$.

\item \textbf{Resizing} is a common operation applied to images, after their acquisition, that changes the dimension of the image. To evaluate the influence of resizing, we set processing parameter values as $\cP = \{\text{1x}, \text{0.5x}, \text{2x}\}$ that define original image, image where both width and height are halved, and image where both width and height are doubled, respectively. We use bilinear interpolation.

\item \textbf{Interpolation} defines the interpolation function used during image resizing. To evaluate the influence of interpolation function, we set processing parameter values as $\cP = \{\text{bilinear}, \text{bicubic}, \text{lanczos}, \text{box}\}$, and we resize each image by changing its both sides by $r\%$ where $r \sim \text{Uniform}[-20,20]$.~\footnote{Note that the same value of $r$ is applied per image across all different interpolations.}

\item \textbf{Make} refers to the manufacturer of the camera, based on Exif metadata. Based on our setup, our analysis is based on nine manufacturers, namely \textit{Apple}, \textit{Canon}, \textit{EASTMAN KODAK COMPANY}, \textit{FUJIFILM}, \textit{NIKON}, \textit{OLYMPUS OPTICAL CO.,LTD}, \textit{Panasonic}, \textit{SONY}, and \textit{Samsung}.

\item \textbf{Model (all)} refers to the specific camera model used to capture the photo. We study 88 different camera models, shown in \cref{tab:camera_models}.

\begin{figure*}
    \centering
    \input{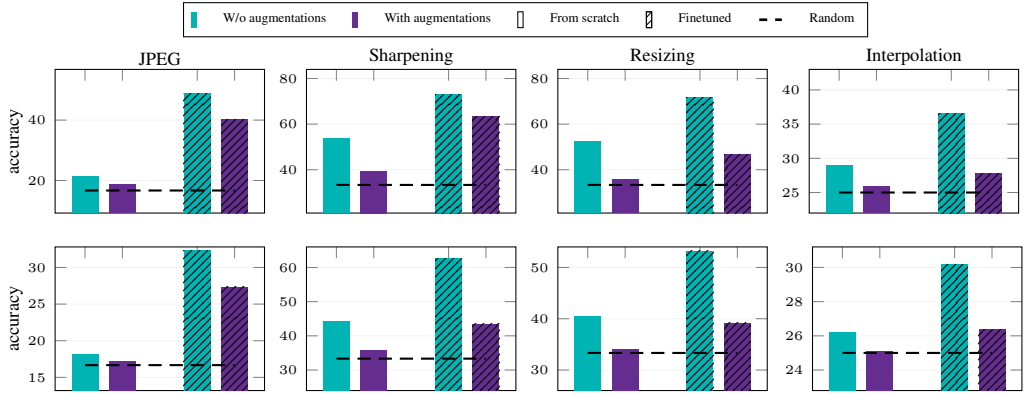}
    \vspace{-25pt}
    \caption{\textbf{Image processing-based label prediction using a CVL model trained with and without augmentations.} Classification accuracy using a linear classifier on embeddings of frozen visual encoders on ImageNet (top) and iNaturalist (bottom) datasets. CVL model trained both from scratch and finetuned starting from a LAION-2B checkpoint.}
    \label{fig:augmentations_processing_prediction}
\end{figure*}

\begin{figure*}
    \centering
    \input{figures/acquisition_plots_aug}
    \vspace{-20pt}
    \caption{\textbf{Image acquisition-based label prediction using a CVL model trained with and without augmentations.} Classification accuracy using a linear classifier on embeddings of frozen visual encoders with images masked at 95\% on the FlickrExif dataset. CVL model trained both from scratch and finetuned starting from a LAION-2B checkpoint.}
    \label{fig:linear_probing_accuracies_aug}
\end{figure*}

\begin{figure*}
    \centering
    \input{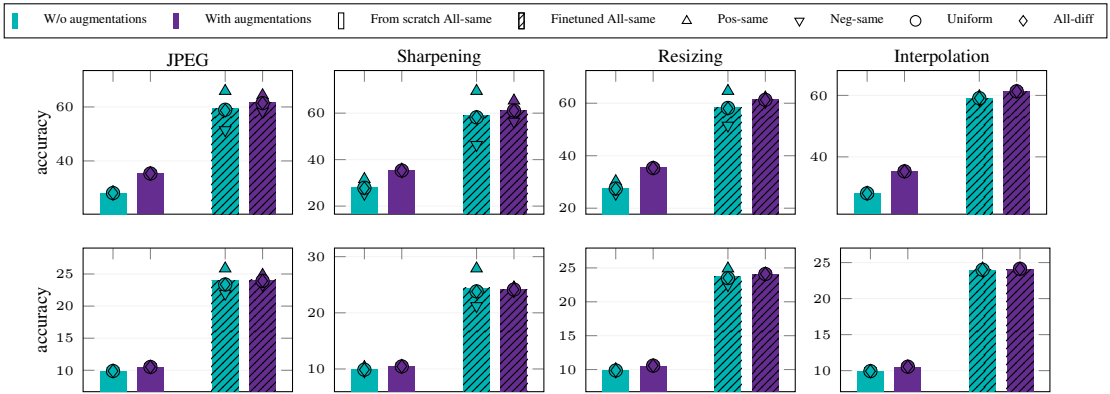}
    \vspace{-20pt}
    \caption{\textbf{Impact of image processing parameters on semantics for a CVL model trained with and without augmentations.} Semantic label prediction accuracy on ImageNet (top) and iNaturalist (bottom) datasets in five different setups. \textit{All-same (baseline)}: all training and test images share the same processing-based metadata label. \textit{All-diff}: training images have the same metadata label, which is different than that of the test image. \textit{Pos-same}: training images that are semantically positive to the test image have the same metadata label as the test image. \textit{Neg-same}: training images that are semantically negative to the test image have the same metadata label as the test image. \textit{Uniform}: the metadata labels are uniformly assigned to the training images. \textit{Pos-same} and \textit{neg-same} settings require an artificially created experiment where a different training set is used per test image. CVL model trained both from scratch and finetuned starting from a LAION-2B checkpoint. Shown for k = 10 and k = 1 for the kNN classifier for ImageNet and iNaturalist, respectively.}
    \label{fig:augmentations_processing_influence}
\end{figure*}

\begin{table*}[t]
\footnotesize
\centering
\vspace{-5pt}
\caption{All 88 camera models studied when analyzing the acquisition attribute model (all). Names are presented as they were found in the Exif metadata. \underline{Underlined} models refer to smartphones analyzed under the model (smart) parameter.}
\vspace{-8pt}

\setlength{\tabcolsep}{3pt}
\scalebox{0.95}{
\begin{tabular}{p{0.2\linewidth} p{0.3\linewidth} p{0.2\linewidth} p{0.2\linewidth}}
\toprule
CYBERSHOT & Canon EOS DIGITAL REBEL & NIKON D3100 & NIKON D800 \\
Canon EOS 10D & Canon EOS DIGITAL REBEL XT & NIKON D3200 & NIKON D810 \\
Canon EOS 20D & Canon EOS R & NIKON D40 & NIKON D850 \\
Canon EOS 300D DIGITAL & Canon EOS R5 & NIKON D5 & NIKON D90 \\
Canon EOS 30D & Canon EOS R6 & NIKON D50 & NIKON Z 6 \\
Canon EOS 350D DIGITAL & Canon EOS REBEL T3i & NIKON D500 & NIKON Z 6\_2 \\
Canon EOS 40D & Canon EOS Rebel T6 & NIKON D5000 & NIKON Z 9 \\
Canon EOS 450D & Canon EOS-1D X & NIKON D5100 & X-T2 \\
Canon EOS 50D & Canon EOS-1D X Mark II & NIKON D5200 & X-T3 \\
Canon EOS 5D & E-M1MarkII & NIKON D5300 & X-T4 \\
Canon EOS 5D Mark II & E5700 & NIKON D5500 & \underline{iPhone 11} \\
Canon EOS 5D Mark III & E990 & NIKON D5600 & \underline{iPhone 11 Pro Max} \\
Canon EOS 5D Mark IV & ILCE-6000 & NIKON D600 & \underline{iPhone 12 Pro} \\
Canon EOS 600D & ILCE-6400 & NIKON D610 & \underline{iPhone 12 Pro Max} \\
Canon EOS 60D & ILCE-7 & NIKON D70 & \underline{iPhone 13 Pro} \\
Canon EOS 6D & ILCE-7M3 & NIKON D700 & \underline{iPhone 6} \\
Canon EOS 6D Mark II & ILCE-7RM2 & NIKON D7000 & \underline{iPhone 6s} \\
Canon EOS 70D & ILCE-7RM3 & NIKON D7100 & \underline{iPhone 7} \\
Canon EOS 7D & Kodak CLAS Digital Film Scanner / HR200 & NIKON D7200 & \underline{iPhone 7 Plus} \\
Canon EOS 7D Mark II & NIKON D100 & NIKON D750 & \underline{iPhone X} \\
Canon EOS 80D & NIKON D200 & NIKON D7500 & \underline{iPhone XR} \\
Canon EOS 90D & NIKON D300 & NIKON D80 & \underline{iPhone XS} \\
\bottomrule
\end{tabular}
}
\vspace{-12pt}
\label{tab:camera_models}
\end{table*}

\item \textbf{Model (smart)} refers to the specific camera model used to capture the photo, but only among photos captured by smartphones. The 12 classes we study are also shown in~\cref{tab:camera_models}.

\item \textbf{Model (smart vs non-smart)} is a binary parameter that indicates whether the camera used to shoot the photo was a smartphone. When analyzing this parameter, we use a subset of data that was curated to conveniently identify non-smartphones images and smartphone images. The former comprise all images taken with a camera manufactured by Canon, Nikon, Fujifilm, Panasonic, or Olympus; while the latter comprise all images taken with a smartphone manufactured by Apple, Google, Huawei, Xiaomi, or Motorola.

\item \textbf{Exposure} refers to the amount of time that light was allowed to enter the camera while taking the photo. This is a rational number, which in our data ranges from $1/1,000$ seconds to $1/30$ seconds.

\item \textbf{Aperture} refers to size of the opening in the lens and the corresponding amount of light thus allowed to enter the camera while taking the photo. This is measured in f-numbers, calculated as the ratio between the focal length and the diameter of the lens opening. In our experiments, these ratios range from $1.8$ to $11$.

\item \textbf{ISO Speed} is a parameter that measures the camera sensor's sensitivity to light, with higher values leading to higher sensitivity and lower values leading to lower sensitivity. In our data, these numbers range from 50 to 3200.

\item \textbf{Focal Length} describes the distance between the center of the camera's lens and the camera's sensor. This is typically measured in mm. Our data covers focal lengths ranging from $4$ mm to $200$ mm.

\end{itemize}

\section{PairsCams dataset}

Cameras used to collect the PairsCams dataset are shown in Table~\ref{tab:custom_dataset} with the number of images taken by each camera.

\begin{table}[t]
\centering
\footnotesize
\caption{Cameras used during data collection. Each object or scene is captured by two cameras, leading to a total of 1,460 photos from 730 pairs.}

\vspace{-3pt}
\begin{tabular}{@{}llrrr}
\toprule
model & type & year & images \\
\midrule
iPhone XR & smartphone & 2018 & 295 \\
Canon IXY 630 & compact & 2014 & 285 \\
Pixel 4 & smartphone & 2019 & 190 \\
iPhone SE (3rd generation) & smartphone & 2022 & 120 \\
Sony Cyber-shot DSC-WX300 & compact & 2013 & 120 \\
Olympus C-8080 Wide Zoom & compact & 2004 & 100 \\
Casio Exilim EX-FH20 & compact & 2008 & 100 \\
Canon EOS 450D & DSLR & 2008 & 80 \\
Xiaomi Poco X5 Pro  & smartphone & 2023 & 40 \\
iPhone 12 & smartphone & 2020 & 26 \\
iPhone 14 Pro & smartphone & 2022 & 25 \\
Olympus $\mu$700 & compact & 2006 & 25 \\
Nothing Phone (2) & smartphone & 2023 & 20 \\
iPhone 12 Pro & smartphone & 2020 & 14 \\
Motorola Moto G XT1032 & smartphone & 2013 & 10 \\
Nikon Coolpix S200 & compact & 2007 & 10 \\
\midrule
total & & & 1,460 \\
\bottomrule
\end{tabular}
\label{tab:custom_dataset}
\end{table}

\section{Visual encoders}

Our visual encoders are acquired from the following repositories: OpenAI,\footnote{\url{https://github.com/OPENAI}} OpenCLIP,\footnote{\url{https://github.com/mlfoundations/open_clip}} timm,\footnote{\url{https://github.com/huggingface/pytorch-image-models}} and FAIR.\footnote{\url{https://github.com/facebookresearch/moco-v3}}
We follow each encoder's default preprocessing to extract image representations. This typically involves resizing images based on their smaller side, followed by center-cropping to the encoder's input resolution, and normalization of the image tensor.

\begin{figure*}[t]
    \centering
    \input{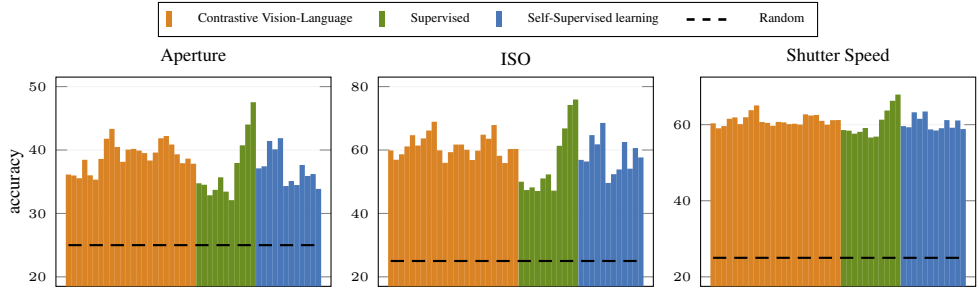}
\pgfplotsset{every tick label/.append style={font=\tiny}}
\pgfplotsset{select coords between index/.style 2 args={
    x filter/.code={
        \ifnum\coordindex<#1\def\pgfmathresult{}\fi
        \ifnum\coordindex>#2\def\pgfmathresult{}\fi
    }
}}
\pgfplotsset{minor grid style={solid,gray,opacity=0.1}}
\pgfplotsset{major grid style={solid,gray,opacity=0.1}}

\pgfplotsset{
    ybar legend/.style={
        legend image code/.code={
            \fill[#1] (0pt,-3pt) rectangle (2pt,3pt);
        }
    }
}
{
\setlength{\tabcolsep}{0.0pt}
\tikzsetnextfilename{imagenet_es_acquisition_plots_1}
\begin{tikzpicture}
    \begin{axis}[
        hide axis,
        scale only axis,
        height=0cm,
        width=0cm,
        legend style={
            draw=black,
            cells={anchor=west},
            legend columns=4,
            column sep=0.8em,
            font=\tiny,
            anchor=north west
        },
        legend entries={
            {Contrastive Vision-Language},
            {Supervised},
            {Self-Supervised learning},
            {Random}
        }
    ]
        \addlegendimage{ybar, fill=vlmcolor, draw=none, ybar legend}
        \addlegendimage{ybar, fill=supervisedcolor, draw=none, ybar legend}
        \addlegendimage{ybar, fill=sslcolor, draw=none, ybar legend}
        \addlegendimage{color=black, line width=1, dashed}
        
        \addplot[draw=none] coordinates {(0,0)};
    \end{axis}
\end{tikzpicture}
\begin{tabular}{c@{\nxssp}c@{\nxssp}c@{\nxssp}}
    \tikzsetnextfilename{imagenet_es_acquisition_plots_2}
    \begin{tikzpicture}
        \begin{axis}[
            width=0.3\linewidth,
            height=0.25\linewidth,
            ylabel={accuracy},
            ylabel style={yshift=-5pt},
            label style={font=\scriptsize},
            title={\scriptsize{Aperture}},
            title style={yshift=-5pt},
            ymin=20, ymax=50,
            bar width=2pt,
            symbolic x coords={0,1,2,3,4,5,6,7,8,9,10,11,12,13,14,15,16,17,18,19,20,21,22,23,24,25,26,27,28,29,30,31,32,33,34,35,36,37,38,39,40,41,42,43,44,45,46},
            xtick=\empty, xlabel={},
            ymajorgrids=true,
            enlargelimits=0.05
        ]
    
            \addplot[ybar, fill=vlmcolor, draw=none] table[x=model, y expr=\thisrow{aperture} * 100] {\vlmimagenetesacquisitionsquaresignal};

            \addplot[ybar, fill=supervisedcolor, draw=none] table[x=model, y expr=\thisrow{aperture} * 100] {\supervisedimagenetesacquisitionsquaresignal};

            \addplot[ybar, fill=sslcolor, draw=none] table[x=model, y expr=\thisrow{aperture} * 100] {\sslimagenetesacquisitionsquaresignal};

            \addplot[black, thick, dash pattern=on 5pt off 3pt, mark=none] coordinates {(0, 25) (46, 25)};
        \end{axis}
    \end{tikzpicture}
    &
    \tikzsetnextfilename{imagenet_es_acquisition_plots_3}
    \begin{tikzpicture}
        \begin{axis}[
            width=0.3\linewidth,
            height=0.25\linewidth,
            ylabel style={yshift=-5pt},
            label style={font=\scriptsize},
            title={\scriptsize{ISO}},
            title style={yshift=-5pt},
            ymin=20, ymax=80,
            bar width=2pt,
            symbolic x coords={0,1,2,3,4,5,6,7,8,9,10,11,12,13,14,15,16,17,18,19,20,21,22,23,24,25,26,27,28,29,30,31,32,33,34,35,36,37,38,39,40,41,42,43,44,45,46},
            xtick=\empty, xlabel={},
            ymajorgrids=true,
            enlargelimits=0.05
        ]
    
            \addplot[ybar, fill=vlmcolor, draw=none] table[x=model, y expr=\thisrow{iso} * 100] {\vlmimagenetesacquisitionsquaresignal};

            \addplot[ybar, fill=supervisedcolor, draw=none] table[x=model, y expr=\thisrow{iso} * 100] {\supervisedimagenetesacquisitionsquaresignal};

            \addplot[ybar, fill=sslcolor, draw=none] table[x=model, y expr=\thisrow{iso} * 100] {\sslimagenetesacquisitionsquaresignal};

            \addplot[black, thick, dash pattern=on 5pt off 3pt, mark=none] coordinates {(0, 25) (46, 25)};
        \end{axis}
    \end{tikzpicture}
    &
    \tikzsetnextfilename{imagenet_es_acquisition_plots_4}
    \begin{tikzpicture}
        \begin{axis}[
            width=0.3\linewidth,
            height=0.25\linewidth,
            ylabel style={yshift=-5pt},
            label style={font=\scriptsize},
            title={\scriptsize{Shutter Speed}},
            title style={yshift=-5pt},
            ymin=20, ymax=70,
            bar width=2pt,
            symbolic x coords={0,1,2,3,4,5,6,7,8,9,10,11,12,13,14,15,16,17,18,19,20,21,22,23,24,25,26,27,28,29,30,31,32,33,34,35,36,37,38,39,40,41,42,43,44,45,46},
            xtick=\empty, xlabel={},
            ymajorgrids=true,
            enlargelimits=0.05
        ]
    
            \addplot[ybar, fill=vlmcolor, draw=none] table[x=model, y expr=\thisrow{shutterspeed} * 100] {\vlmimagenetesacquisitionsquaresignal};

            \addplot[ybar, fill=supervisedcolor, draw=none] table[x=model, y expr=\thisrow{shutterspeed} * 100] {\supervisedimagenetesacquisitionsquaresignal};

            \addplot[ybar, fill=sslcolor, draw=none] table[x=model, y expr=\thisrow{shutterspeed} * 100] {\sslimagenetesacquisitionsquaresignal};

            \addplot[black, thick, dash pattern=on 5pt off 3pt, mark=none] coordinates {(0, 25) (46, 25)};
        \end{axis}
    \end{tikzpicture}
\end{tabular}
}
    \caption{\textbf{Image acquisition-based label prediction on ImageNet-ES.} Classification accuracy using a linear classifier on embeddings of frozen visual encoders with no masking. Ordering is according to \cref{tab:models}.}
    \label{fig:imagenet_eslinear_probing_accuracies}
\end{figure*}

\section{Additional results on ImageNet-ES}

An existing dataset that can be used for acquisition label prediction is ImageNet-ES~\cite{baek2024unexplored}, which contains ImageNet images that have been recaptured under varying acquisition settings, including ISO, shutter speed, aperture, and lighting conditions. Originally designed for out-of-distribution detection, the dataset features disjoint test and training labels. Thus, we randomly split the provided validation set into our own training and test sets using a 9:1 ratio. We follow the hyperparameter tuning and training protocol  described in Sec. 3, using $12.5\%$ of the training set for validation. Each image is annotated with metadata labels according to their aperture, ISO, and shutter speed labels, formulating a four-class classification task for each attribute.

\cref{fig:imagenet_eslinear_probing_accuracies} presents the performance of classifiers trained on frozen embeddings for the prediction of each attribute. Models across all categories achieve classification accuracy well above random chance. We attribute this to the broad range of values of the acquisition parameters used to create ImageNet-ES. For example, the ISO values of ImageNet-ES's validation set ranges from $200$ to $12,800$, while the corresponding values in FlickrExif only range from $50$ to $3,200$. This wider range may result in more visually distinguishable cases compared to those in FlickrExif.

\section{Effect of masking}
\begin{figure*}[t]
    \centering
    \input{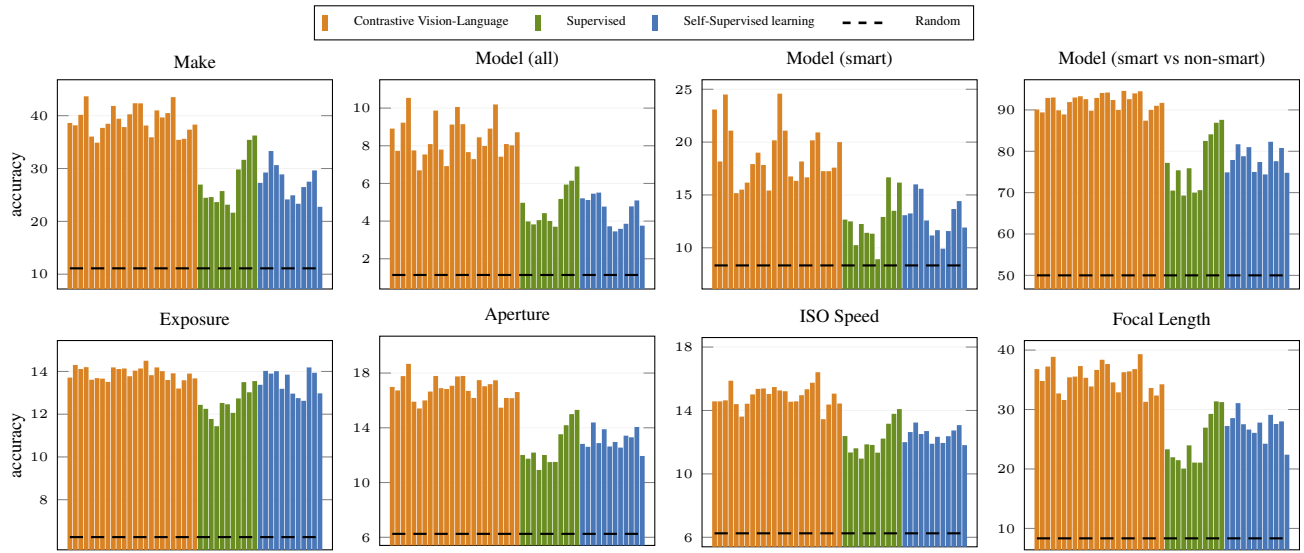}
    \caption{\textbf{Image acquisition-based label prediction on FlickrExif without masking.} Classification accuracy using a linear classifier. Ordering is according to \cref{tab:models}.}
    \label{fig:linear_probing_accuracies}
\end{figure*}

In this section, we assess the impact of the masking applied for the prediction of acquisition labels in Sec. 3.2. \cref{fig:linear_probing_accuracies} and \cref{fig:linear_probing_accuracies_75} show the classification accuracy on FlickrExif with $0\%$ and $75\%$ masking, respectively. Comparing the two figures with Fig. 6, we observe that retaining semantic information in the input images makes it easier to identify acquisition labels, leading to higher classification accuracy. This suggests potential correlations between acquisition labels and semantic content, which can be exploited by the models to achieve a better performance.

\begin{figure*}[t]
    \centering
    \input{figures/acquisition_plots_75_v2}
    \caption{\textbf{Image acquisition-based label prediction on FlickrExif with a $75\%$ masking ratio.} Classification accuracy using a linear classifier. Ordering is according to \cref{tab:models}.}
    \label{fig:linear_probing_accuracies_75}
\end{figure*}

\begin{table*}[t]
\centering
\setlength{\tabcolsep}{10pt}
\footnotesize
\caption{List of all visual encoders used with their characteristics.}

\vspace{-3pt}
\begin{tabular}{@{}r l l l l r r r l}
\toprule
id & model & variant & arch & class & dim & resolution & params (M) & train dataset \\
\midrule
1 & CLIP & ViT-B/16 & Transformer & CVL & 512 & 224 & 88 & WIT\\
2 & CLIP & ViT-B/32 & Transformer & CVL & 512 & 224 & 86 & WIT\\
3 & CLIP & ViT-L/14 & Transformer & CVL & 768 & 224 & 304 & WIT\\
4 & CLIP & ViT-L/14@336 & Transformer & CVL & 768 & 336 & 304 & WIT\\
5 & CLIP & RN50 & CNN & CVL & 1024 & 224 & 38 & WIT\\
6 & CLIP & RN101 & CNN & CVL & 512 & 224 & 56 & WIT\\
7 & CLIP & RN50×4 & CNN & CVL & 640 & 288 & 87 & WIT\\
8 & CLIP & RN50×16 & CNN & CVL & 768 & 384 & 167 & WIT\\
9 & CLIP & RN50×64 & CNN & CVL & 1024 & 448 & 420 & WIT\\
10 & OpenCLIP & ViT-B/16 & Transformer & CVL & 512 & 224 & 86 & LAION-2B\\
11 & OpenCLIP & ViT-B/32 & Transformer & CVL & 512 & 224 & 87 & LAION-2B\\
12 & OpenCLIP & ViT-L/14 & Transformer & CVL & 768 & 224 & 303 & LAION-2B\\
13 & OpenCLIP & ViT-H/14 & Transformer & CVL & 1024 & 224 & 632 & LAION-2B\\
14 & OpenCLIP & ViT-g/14 & Transformer & CVL & 1024 & 224 & 1012 & LAION-2B\\
15 & OpenCLIP & ViT-B/16 & Transformer & CVL & 512 & 224 & 86 & DataComp-1B\\
16 & OpenCLIP & ViT-B/32 & Transformer & CVL & 512 & 256 & 87 & DataComp-1B\\
17 & OpenCLIP & ViT-L/14 & Transformer & CVL & 768 & 224 & 303 & DataComp-1B\\
18 & OpenCLIP & ConvNeXt-B & CNN & CVL & 640 & 256 & 88 & LAION-2B\\
19 & OpenCLIP & ConvNeXt-L & CNN & CVL & 768 & 320 & 199 & LAION-2B\\
20 & OpenCLIP & ConvNeXt-XXL & CNN & CVL & 1024 & 256 & 846 & LAION-2B\\
21 & SigLIP & ViT-B/16 & Transformer & CVL & 768 & 256 & 93 & WebLI\\
22 & SigLIP & ViT-L/16 & Transformer & CVL & 1024 & 256 & 316 & WebLI\\
23 & SigLIP2 & ViT-B/16 & Transformer & CVL & 768 & 256 & 93 & WebLI\\
24 & SigLIP2 & ViT-L/16 & Transformer & CVL & 1024 & 256 & 316 & WebLI\\
\midrule
25 & ViT & ViT-B/16 & Transformer & SUP & 768 & 224 & 86 & ImageNet-21k\\
26 & ViT & ViT-B/32 & Transformer & SUP & 768 & 224 & 86 & ImageNet-21k\\
27 & ViT & ViT-L/16 & Transformer & SUP & 1024 & 224 & 307 & ImageNet-21k\\
28 & ViT & ViT-L/32 & Transformer & SUP & 1024 & 224 & 307 & ImageNet-21k\\
29 & ViT & ViT-H/14 & Transformer & SUP & 1280 & 224 & 632 & ImageNet-21k\\
30 & ResNet & RN50 & CNN & SUP & 2048 & 224 & 26 & ImageNet-1k\\
31 & ResNet & RN101 & CNN & SUP & 2048 & 224 & 45 & ImageNet-1k\\
32 & ConvNeXt & ConvNeXt-T & CNN & SUP & 768 & 384 & 50 & ImageNet-21k\\
33 & ConvNeXt & ConvNeXt-B & CNN & SUP & 1024 & 384 & 89 & ImageNet-21k\\
34 & ConvNeXt & ConvNeXt-L & CNN & SUP & 1536 & 384 & 198 & ImageNet-21k\\
35 & ConvNeXt & ConvNeXt-XL & CNN & SUP & 2048 & 384 & 350 & ImageNet-21k\\
\midrule
36 & DINO & ViT-S/16 & Transformer & SSL & 384 & 224 & 21 & ImageNet-1k\\
37 & DINO & ViT-S/8 & Transformer & SSL & 384 & 224 & 21 & ImageNet-1k\\
38 & DINO & ViT-B/16 & Transformer & SSL & 768 & 224 & 85 & ImageNet-1k\\
39 & DINO & ViT-B/8 & Transformer & SSL & 768 & 224 & 85 & ImageNet-1k\\
40 & DINO & RN50 & CNN & SSL & 2048 & 224 & 23 & ImageNet-1k\\
41 & DINOv2 & ViT-S/14 reg & Transformer & SSL & 384 & 224 & 21 &  LVD-142M\\
42 & DINOv2 & ViT-B/14 reg & Transformer & SSL & 768 & 224 & 86 &  LVD-142M\\
43 & DINOv2 & ViT-L/14 reg & Transformer & SSL & 1024 & 224 & 300 &  LVD-142M\\
44 & DINOv2 & ViT-g/14 reg & Transformer & SSL & 1536 & 224 & 1100 &  LVD-142M\\
45 & MoCo v3 & ViT-S & Transformer & SSL & 384 & 224 & 22 & ImageNet-1k\\
46 & MoCo v3 & ViT-B & Transformer & SSL & 768 & 224 & 86 & ImageNet-1k\\
47 & MoCo v3 & RN50 & CNN & SSL & 2048 & 224 & 26 & ImageNet-1k\\
\bottomrule
\end{tabular}
\label{tab:models}
\end{table*}

\end{document}